\documentclass[british,english]{article}
\usepackage[T1]{fontenc}
\usepackage[latin9]{inputenc}
\usepackage{xcolor}
\usepackage{babel}
\usepackage{textcomp}
\usepackage{mathrsfs}
\usepackage{amstext}
\usepackage{graphicx}
\usepackage[unicode=true,pdfusetitle,
 bookmarks=true,bookmarksnumbered=false,bookmarksopen=false,
 breaklinks=false,pdfborder={0 0 1},backref=false,colorlinks=false]
 {hyperref}

\makeatletter

\providecommand{\tabularnewline}{\\}

\newcommand{\lyxaddress}[1]{
	\par {\raggedright #1
	\vspace{1.4em}
	\noindent\par}
}

\@ifundefined{date}{}{\date{}}
\@ifundefined{showcaptionsetup}{}{%
 \PassOptionsToPackage{caption=false}{subfig}}
\usepackage{subfig}
\makeatother

\begin{document}
\title{Dynamic Fusion Network for RGBT Tracking}
\author{Jingchao Peng, Haitao Zhao, Zhengwei Hu}
\maketitle

\lyxaddress{\begin{center}
Automation Department, School of Information Science and Engineering,
\\
East China University of Science and Technology, Shanghai, China
\par\end{center}}
\begin{abstract}
For both visible and infrared images have their own advantages and
disadvantages, RGBT tracking has attracted more and more attention.
The key points of RGBT tracking lie in feature extraction and feature
fusion of visible and infrared images. Current RGBT tracking methods
mostly pay attention to both \textit{individual features} (features
extracted from images of a single camera) and \textit{common features}
(features extracted and fused from an RGB camera and a thermal camera),
while pay less attention to the different and dynamic contributions
of individual features and common features for different sequences
of registered image pairs. This paper proposes a novel RGBT tracking
method, called Dynamic Fusion Network (DFNet), which adopts a two-stream
structure, in which two non-shared convolution kernels are employed
in each layer to extract individual features. Besides, DFNet has shared
convolution kernels for each layer to extract common features. Non-shared
convolution kernels and shared convolution kernels are adaptively
weighted and summed according to different image pairs, so that DFNet
can deal with different contributions for different sequences. DFNet
has a fast speed, which is 28.658 FPS. The experimental results show
that when DFNet only increases the Mult-Adds of 0.02\% than the non-shared-convolution-kernel-based
fusion method, Precision Rate (PR) and Success Rate (SR) reach 88.1\%
and 71.9\% respectively.
\end{abstract}

\section{Introduction \label{sec:Introduction}}

Object tracking is a popular computer vision task, whose purpose is
to continuously track the position of the object in the subsequent
frames when given in the first frame. Tracking in a complex visual
scenery, including rain, smoke, or night, is one of the most difficult
computer vision tasks \cite{ref1,ref2}, especially for visible-image-based
tracking \cite{ref2,ref3}. However, infrared sensors can work around
the clock, infrared has a strong ability to penetrate smoke, which
can supplement the deficiencies of visible images in bad visual conditions
\cite{ref4,ref5,ref6,ref7}. Therefore, RGBT tracking has attracted
more and more attention. 

Since 2018, due to the powerful learning ability, Deep Learning (DL)
models, especially Convolutional Neural Networks (CNN), are widely
used to address RGBT tracking \cite{pixel1,pixel2,ref28,ref9,ref11,ref10,ref15,deci1}.
DL-based tracking methods have demonstrated their capabilities over
traditional fusion tracking methods \cite{tri1,tri2,tri3,tri4} or
other tracking methods (e.g., sparse representation-based methods
\cite{sr1,sr2,sr3}, and graph-based methods \cite{g1,g2,g3,g4}).
The advantage of DL-based tracking methods is their ability to learn
more effective and robust features than hand-crafted features \cite{deepfeatures1,deepfeatures2,ref3}.
DL-based tracking methods can be divided into pixel-level \cite{pixel1,pixel2,ref28},
feature-level \cite{ref9,ref11,ref10,ref15}, and decision-level \cite{deci1}
fusion tracking. Compared with the pixel-level fusion method, the
feature-level fusion method has lower requirements for image registration
and can tolerate a certain amount of noise \cite{ref11,ref10}. Compared
with the decision-level fusion method, it has lower computational
complexity and faster speed \cite{SIAMIVFN,ref3}. Recently research
works of DL-based RGBT tracking mainly focus on feature-level fusion
\cite{ref3}.

Due to visible light reflection and infrared radiation have different
imaging properties, visible and infrared images have different \textit{individual
features} \cite{ref8}, which can be used to track objects based on
single-modal images. In visible-image-based tracking, objects can
be distinguished through rich textures and different colors. While
in infrared-image-based tracking, objects can be distinguished by
high-contrast light-dark changes that reflect the heat of the object.
In order to fully utilize the individual features from the two different
modalities, feature-level fusion methods are often adopted, in which
two Convolutional Neural Networks (CNNs) were often employed to handle
visible and infrared images, respectively. For example, Zhang et al.
\cite{ref9} utilized two different CNNs to respectively extract individual
features from visible and infrared images. In their work the visible
and infrared features were concatenated and sent to follow layers
for tracking the object. ConvNet \cite{ref10} and SiamFT \cite{ref11}
employ fusion sub-networks to select discriminative features after
extracting the individual features. DSiamMFT \cite{ref12} and FANet
\cite{ref13} focus on multi-layer feature fusion to achieve more
effective hierarchical feature aggregation. For simplicity, this paper
denotes the basic feature-level fusion method without any bells and
whistles, which only uses two different CNNs to respectively handle
visible and infrared images, as non-shared-convolution-kernel-based
fusion method.

In addition to individual features, since visible and infrared images
are shot in the same scene and are used to track the same object,
there are \textit{common features} in the two modalities \cite{ref14}.
Common features reflect the size, location, contour, and so on, which
are also important information in object tracking \cite{ref16}. When
individual features are not enough to achieve good tracking performances,
it is necessary to use common features such as the semantics of the
object, and other characteristics of the object at the corresponding
position of the visible and infrared images for object tracking. Both
MANet \cite{ref15} , CAT \cite{ref8} , IVFuseNet \cite{ref16},
and SiamIVFN \cite{SIAMIVFN} use a shared convolution kernel to extract
common features. Their experimental results show that the shared-convolution-kernel-based
fusion methods can extract common features that are more informative
than the non-shared-convolution-kernel-based fusion method. But in
their networks, the contributions of individual features and common
features are prefixed and have no consideration of adaption to the
registered image pairs captured in different scenes.

\begin{figure}
\begin{centering}
\subfloat[GTOT: \textit{OccCar}]{\begin{centering}
\includegraphics[scale=0.4]{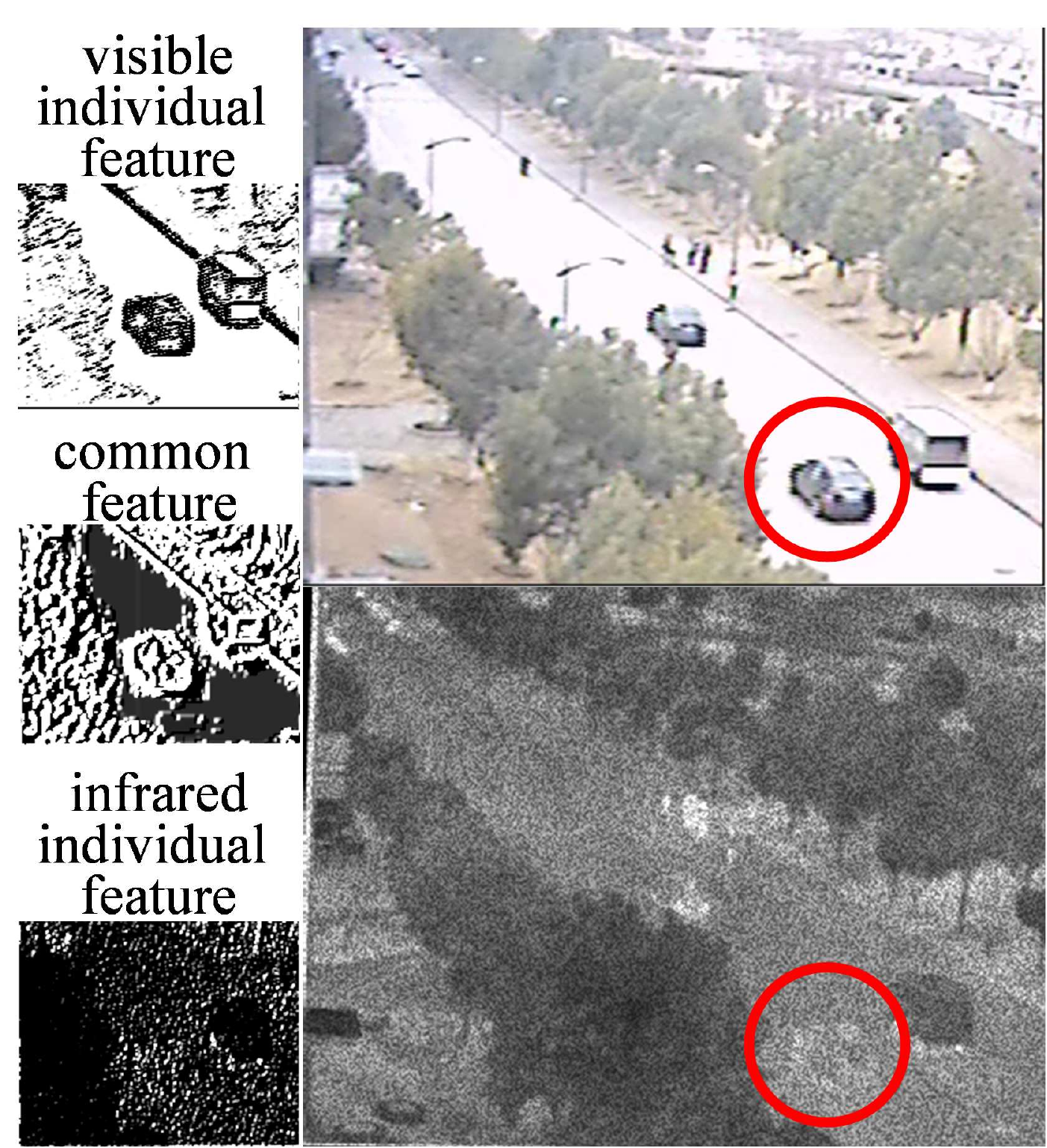}
\par\end{centering}
}\subfloat[GTOT: \textit{OccBike}]{\begin{centering}
\includegraphics[scale=0.4]{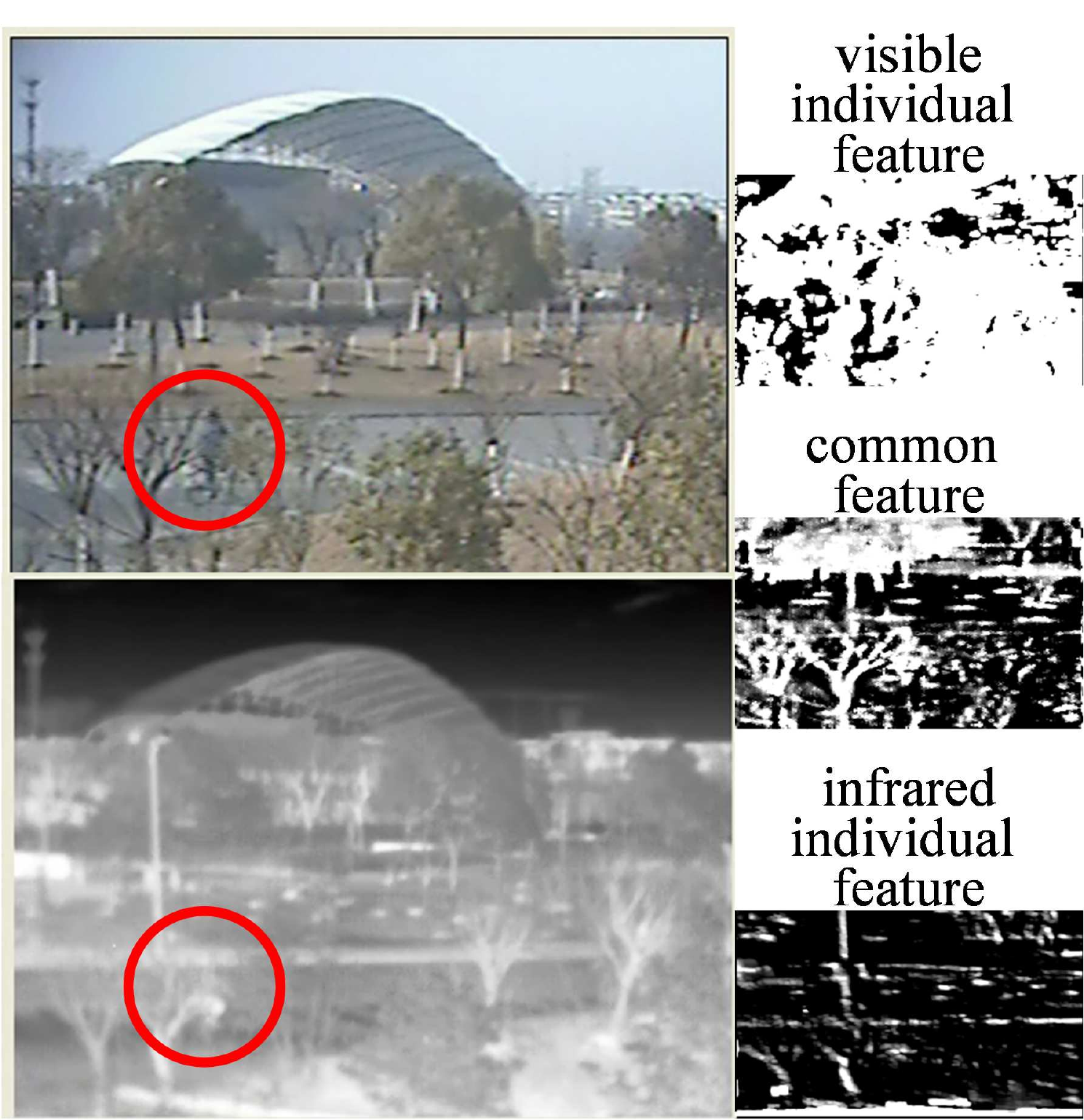}
\par\end{centering}
}
\par\end{centering}
\caption{\label{fig:1}Two registered image pairs with different contributions
of individual features and common features. The two images in the
first row are visible images and the other two images in the second
row are the corresponding registered infrared images. The images in
(a) are from GTOT: \textit{OccCar}. The visible image is clear but
the infrared image is not clear because of noise, so the tracking
task should pay more attention to the individual features of the visible
image. The images in (b) are from GTOT: \textit{OccBike}. the target
is not easy to detect in both visible and infrared images due to the
messy background. The tracking task should pay more attention to the
common features.}
\end{figure}

However, the contributions of individual features and common features
are not always fixed. Visible images are greatly affected by the illumination
and prone to overexposure or underexposure. Infrared images are easy
to be interfered with by the external scenery and internal systems
and prone to noise \cite{addnoise1,addnoise2}. In other words, the
reliability of different modalities is not always fixed. For different
reliable modalities, individual features and common features contribute
to different degrees. When one modality is reliable, the individual
features of the reliable modality contribute more, as shown in Figure
\ref{fig:1} (a). When it is impossible to track based on single-modal
images, more attention needs to be paid to common features, as shown
in Figure \ref{fig:1} (b). Therefore, the tracker needs to adaptively
calculate different contributions of individual features and common
features in different scenes.

To solve the performance limitation of the network in changing scenes,
dynamic convolution has natural advantages. The concept of dynamic
convolution (e.g., CondConv \cite{ref18}, Dynamic Convolution \cite{ref17},
and WeightNet \cite{ref17}) usually adopts the method of attention
over convolution kernels. Dynamic convolution has been applied in
scene segmentation, scene synthesizing, image inpainting, biomedical
imaging, and so on \cite{adddc1,adddc2,adddc3,adddc4}. Due to aggregating
multiple convolution kernels adapted to each input, dynamic convolution
has more representation power without increasing the width and depth
of the network. The aggregation of multiple convolution kernels in
convolution kernel space makes it possible to make full use of multiple
convolution kernels only by one convolution operation. Therefore,
dynamic convolution is computationally efficient. But dynamic convolution
is designed for integration into existing CNN architectures, cannot
aggregate individual features and common features in fusion tasks.

Motivated by the above analysis, we propose a novel RGBT tracking
method called dynamic fusion network (DFNet). DFNet adopts a two-stream
structure, which has non-shared convolution kernels to extract individual
features. One CNN is utilized to extract features from infrared images,
and the other one is for handle visible images. Besides, DFNet has
shared convolution kernels to extract common features. DFNet adaptively
merges the shared convolution kernels and the non-shared convolution
kernels in convolution kernel space through the dynamic convolution.
To satisfy strict latency requirements for object tracking, DFNet
only needs two convolution operations in each layer to extract the
individual and common features of visible and infrared images. Since
the weights of shared and non-shared convolution kernels are dynamically
computed, it can adaptively calculate the contributions of individual
features and common features to different scenes.

Specifically, the proposed method has the following advantages: 
\begin{enumerate}
\item DFNet has shared kernels and non-shared kernels that separately extract
the common features and individual features. DFNet has a strong representation
power. 
\item DFNet adaptively calculates the contributions of individual features
and common features according to different registered image pairs. 
\item By fusing multiple kernels in convolution kernel space, DFNet boosts
the PR/SR by 1.1\%/0.9\% with only 0.02\% additional Mult-Adds. 
\end{enumerate}

\section{Related Work }

\begin{figure}
\subfloat[Conv Layer]{\begin{centering}
\includegraphics[scale=0.25]{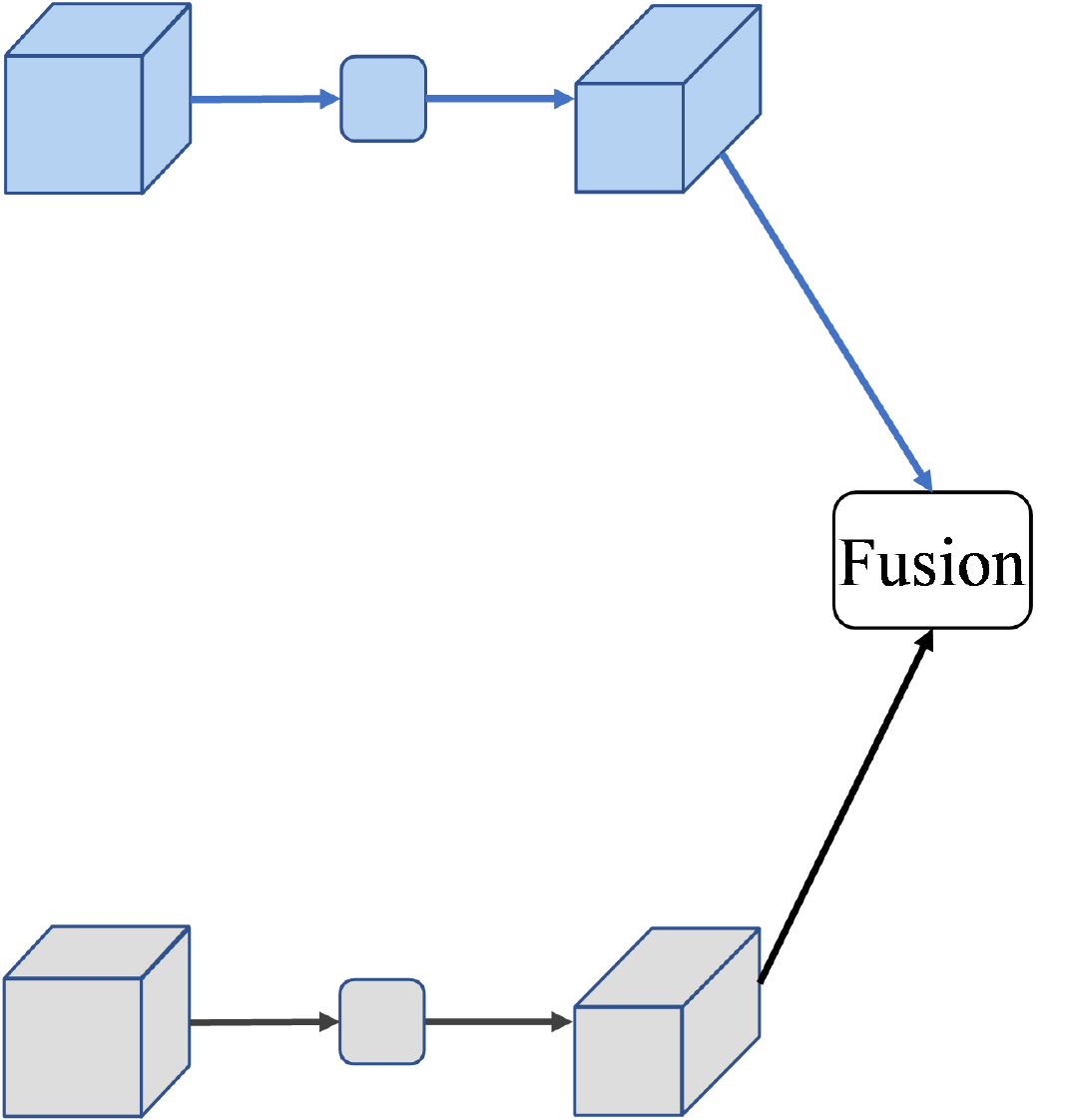}
\par\end{centering}
}\subfloat[MA Layer]{\begin{centering}
\includegraphics[scale=0.25]{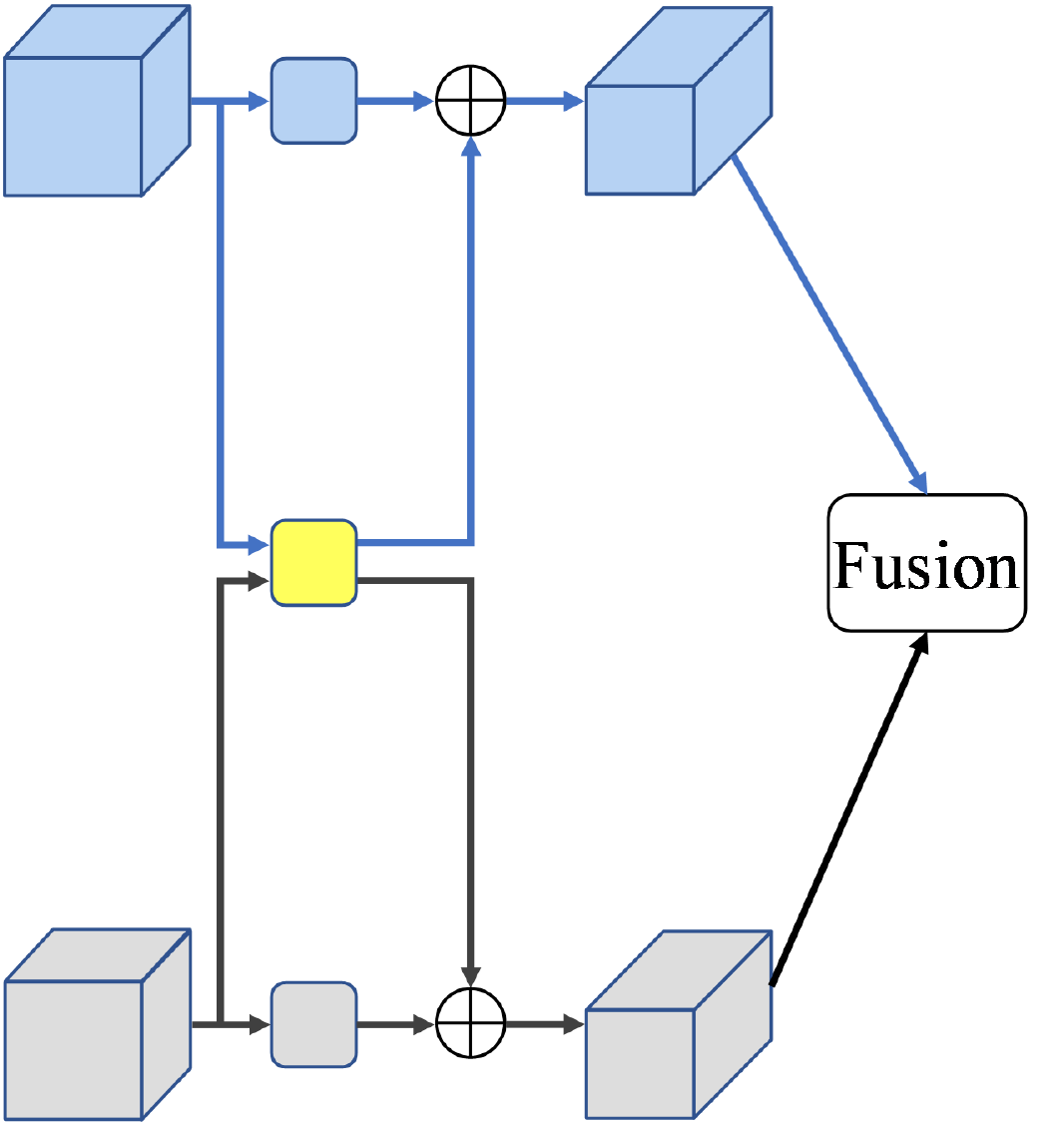}
\par\end{centering}
}\subfloat[IVFuse Layer]{\begin{centering}
\includegraphics[scale=0.25]{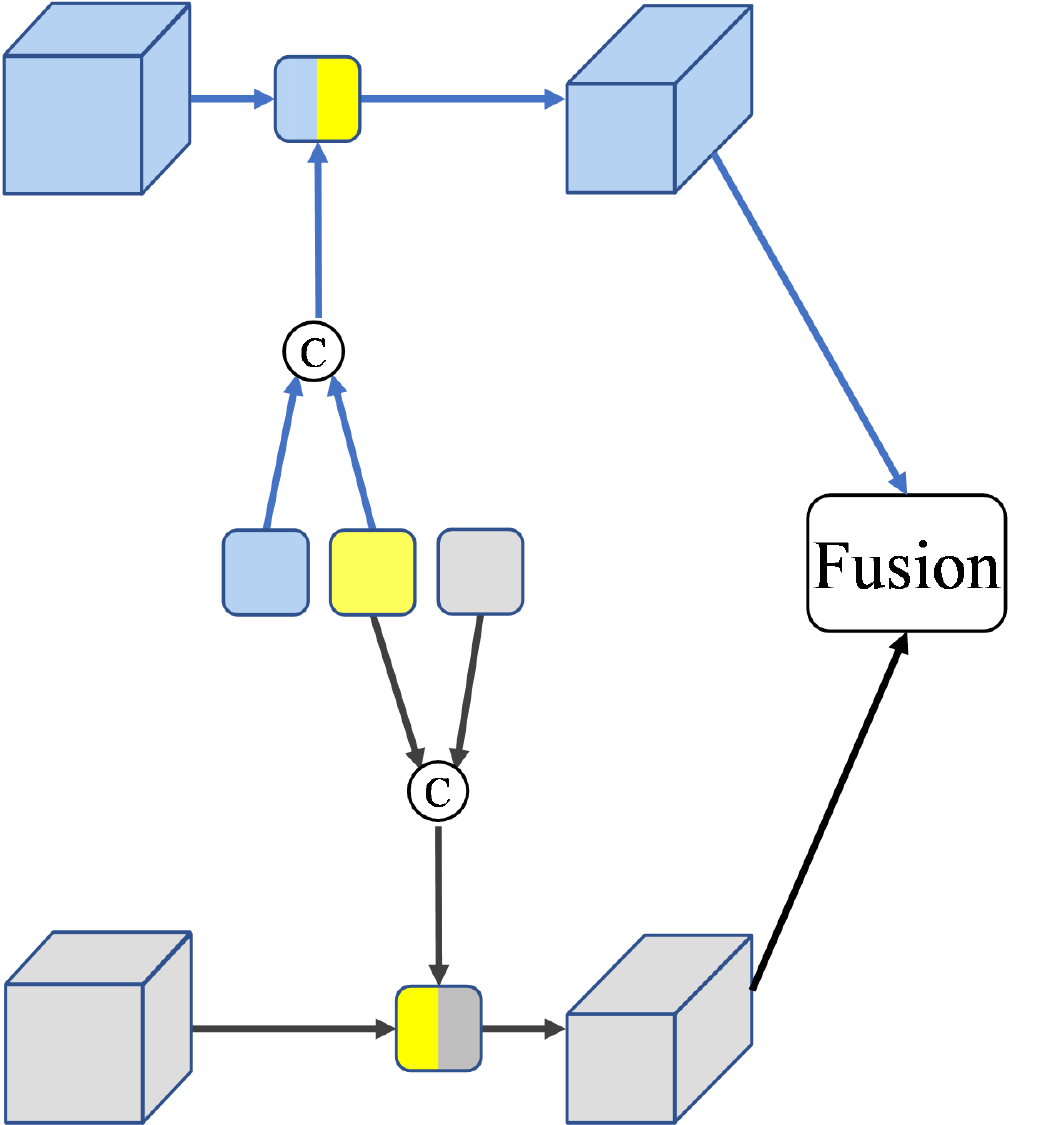}
\par\end{centering}
}\subfloat[DF Layer (ours)]{\begin{centering}
\includegraphics[scale=0.25]{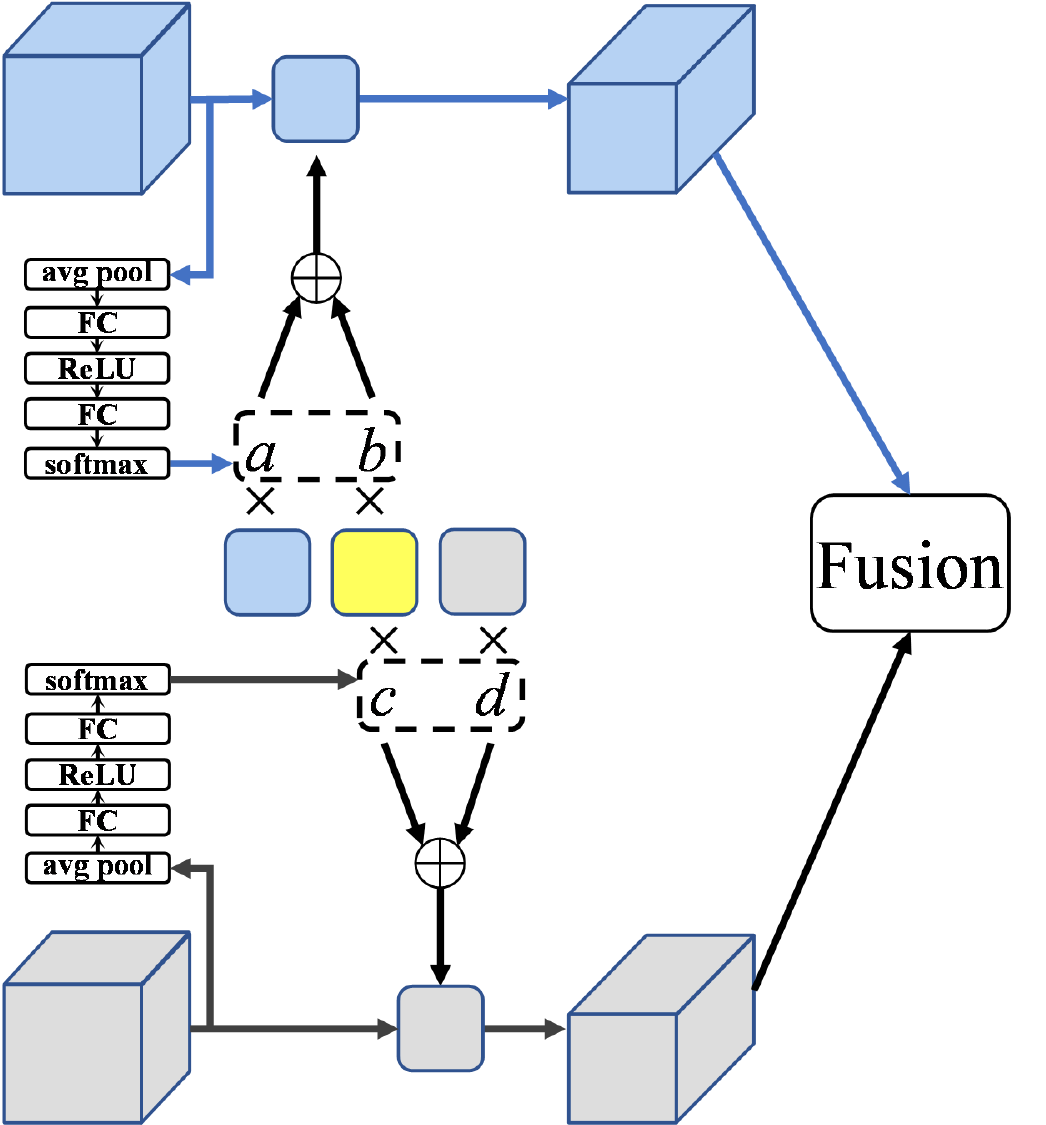}
\par\end{centering}
}

\caption{\label{fig:2}Simplified feature extraction layer diagrams of (a)
ConvNet. The blue and gray branches respectively handle visible and
infrared images. (b) MANet. The yellow block represents the shared
convolution kernel which is used to extract common features. (c) IVFuseNet.
The shared and the non-shared convolution kernels are concatenated
to form a new convolution kernel for feature extraction. (d) DFNet
(ours) The shared and the non-shared convolution kernels are weighted
and summed to form a new convolution kernel for feature extraction.
The weights are calculated based on the input.}
\end{figure}

Section \ref{sec:Introduction} overviews DL-based RGBT tracking.
This section focuses on three most related works to ours: ConvNet
\cite{ref10}, MANet \cite{ref15}, IVFuseNet \cite{ref16}. Their
simplified feature extraction layer diagrams are shown in Figure \ref{fig:2}.

\begin{table}
\caption{\label{tab:1}Expressions of different fusion methods. $W_{RGB}$
and $W_{T}$ represent the non-shared convolution kernel for RGB features
and the convolution kernel for thermal features respectively. $W_{share}$
represents the shared convolution kernel. $F_{RGB}$ and $F_{T}$
represent the input of visible and infrared branch respectively. $\ast$
represents convolution operation. $\sigma\left(\cdot\right)$ represents
activation function.}

\selectlanguage{british}%
\centering{}\resizebox{\textwidth}{!}{
\begin{tabular}{cc}
\hline 
\selectlanguage{english}%
Fusion method\selectlanguage{british}%
 & \selectlanguage{english}%
Expression\selectlanguage{british}%
\tabularnewline
\hline 
\selectlanguage{english}%
ConvNet\selectlanguage{british}%
 & \selectlanguage{english}%
{\small{}$\begin{array}{c}
\sigma\left(\left[\begin{array}{cc}
W_{RGB} & W_{T}\end{array}\right]\ast\left[\begin{array}{cc}
F_{RGB}\\
 & F_{T}
\end{array}\right]\right)\\
=\left[\begin{array}{cc}
\sigma\left(W_{RGB}\ast F_{RGB}\right) & \sigma\left(W_{T}\ast F_{T}\right)\end{array}\right]
\end{array}$}\selectlanguage{british}%
\tabularnewline
\selectlanguage{english}%
MANet\selectlanguage{british}%
 & \selectlanguage{english}%
{\small{}$\begin{array}{c}
\left[\begin{array}{ccc}
\sigma W_{RGB} & \sigma W_{share} & \sigma W_{T}\end{array}\right]\ast\left(\begin{array}{cc}
\left[\begin{array}{cc}
1\\
1 & 1\\
 & 1
\end{array}\right] & \left[\begin{array}{cc}
F_{RGB}\\
 & F_{T}
\end{array}\right]\end{array}\right)\\
=\left[\begin{array}{cc}
\sigma\left(W_{RGB}\ast F_{RGB}\right)+\sigma\left(W_{share}\ast F_{RGB}\right) & \sigma\left(W_{share}\ast F_{T}\right)+\sigma\left(W_{T}\ast F_{T}\right)\end{array}\right]
\end{array}$}\selectlanguage{british}%
\tabularnewline
\selectlanguage{english}%
IVFuseNet\selectlanguage{british}%
 & \selectlanguage{english}%
{\small{}$\begin{array}{c}
\sigma\left(\left(\left[\begin{array}{ccc}
W_{RGB} & W_{share} & W_{T}\end{array}\right]\left[\begin{array}{cccc}
1\\
 & 1 & 1\\
 &  &  & 1
\end{array}\right]\right)\ast\left[\begin{array}{cc}
F_{RGB}\\
 & F_{T}
\end{array}\right]\right)\\
=\left[\begin{array}{cc}
\sigma\left(\left[\begin{array}{cc}
W_{RGB} & W_{share}\end{array}\right]\ast F_{RGB}\right) & \sigma\left(\left[\begin{array}{cc}
W_{share} & W_{T}\end{array}\right]\ast F_{T}\right)\end{array}\right]
\end{array}$}\selectlanguage{british}%
\tabularnewline
\selectlanguage{english}%
DFNet\selectlanguage{british}%
 & \selectlanguage{english}%
{\small{}$\begin{array}{c}
\sigma\left(\left(\left[\begin{array}{ccc}
W_{RGB} & W_{share} & W_{T}\end{array}\right]\left[\begin{array}{cc}
a\\
b & c\\
 & d
\end{array}\right]\right)\ast\left[\begin{array}{cc}
F_{RGB}\\
 & F_{T}
\end{array}\right]\right)\\
=\left[\begin{array}{cc}
\sigma\left(\left(aW_{RGB}+bW_{share}\right)\ast F_{RGB}\right) & \sigma\left(\left(cW_{share}+dW_{T}\right)\ast F_{T}\right)\end{array}\right]
\end{array}$}\selectlanguage{british}%
\tabularnewline
\hline 
\end{tabular}}\selectlanguage{english}%
\end{table}

\subsection{ConvNet }

ConvNet \cite{ref10} uses different convolutional networks to extract
the individual features of visible and infrared images and then fuses
them, its feature extraction layer can be expressed as: 

{\small{}
\begin{equation}
\begin{array}{c}
\sigma\left(\left[\begin{array}{cc}
W_{RGB} & W_{T}\end{array}\right]\ast\left[\begin{array}{cc}
F_{RGB}\\
 & F_{T}
\end{array}\right]\right)\\
=\left[\begin{array}{cc}
\sigma\left(W_{RGB}\ast F_{RGB}\right) & \sigma\left(W_{T}\ast F_{T}\right)\end{array}\right]
\end{array}
\end{equation}
}where $W_{RGB}$ and $W_{T}$ represent the convolution kernel for
RGB features and the convolution kernel for thermal features respectively,
$F_{RGB}$ and $F_{T}$ represent the input of visible and infrared
branch respectively, $\ast$ represents convolution operation, $\sigma\left(\cdot\right)$
represents activation function, such as ReLU.

In ConvNet, different convolution kernels are used to extract individual
features from visible and infrared images. Then these two features
are fused and sent to domain-specific layers for binary classification
and identification of the target. Besides, ConvNet designs a fusion
sub-network, which adaptively fuses two individual features to removing
redundant noise. The feature extraction process performed two convolution
operations in one layer, therefore the speed of ConvNet is fast. ConvNet
focuses on individual features but does not fully consider the common
features.

\subsection{MANet}

Li, et al. \cite{ref15} argue that common features of visible and
infrared images is crucial to the effectiveness of the fusion. Therefore,
MANet employs a shared convolution kernel to extract the common features
of visible and infrared images. The feature extraction layer of MANet
can be expressed as:

{\footnotesize{}
\begin{equation}
\begin{array}{c}
\left[\begin{array}{ccc}
\sigma W_{RGB} & \sigma W_{share} & \sigma W_{T}\end{array}\right]\ast\left(\begin{array}{cc}
\left[\begin{array}{cc}
1\\
1 & 1\\
 & 1
\end{array}\right] & \left[\begin{array}{cc}
F_{RGB}\\
 & F_{T}
\end{array}\right]\end{array}\right)\\
=\left[\begin{array}{cc}
\sigma\left(W_{RGB}\ast F_{RGB}\right)+\sigma\left(W_{share}\ast F_{RGB}\right) & \sigma\left(W_{share}\ast F_{T}\right)+\sigma\left(W_{T}\ast F_{T}\right)\end{array}\right]
\end{array}
\end{equation}
}where $W_{share}$ represents the shared convolution kernel.

Before respective convolution operations, visible and infrared features
must both undergo a shared convolution operation, which uses the same
convolution kernel. It is worth noting that MANet fuses shared and
non-shared features in the feature space. Please note that the activation
function $\sigma\left(\cdot\right)$ is not a linear operation, and
we have:

\begin{equation}
\sigma\left(W_{RGB}\ast F_{RGB}\right)+\sigma\left(W_{share}\ast F_{RGB}\right)\neq\sigma\left(\left(W_{RGB}+W_{share}\right)\ast F_{RGB}\right)
\end{equation}
\begin{equation}
\sigma\left(W_{share}\ast F_{T}\right)+\sigma\left(W_{T}\ast F_{T}\right)\neq\sigma\left(\left(W_{share}+W_{T}\right)\ast F_{T}\right)
\end{equation}
Therefore, four convolution operations are needed. The complexity
of the operation is large, which is not conducive to the real-time
requirements of the tracking task. In addition, the weights of the
shared and non-shared convolution kernel are equal, so that they cannot
be adjusted in real-time in the face of different contributions of
individual features and common features.

\subsection{IVFuseNet}

Unlike the fusion of shared and non-shared features in feature space,
IVFuseNet \cite{ref16} merges the shared and non-shared convolution
kernels in convolution kernel space. IVFuseNet concatenates two small-sized
convolution kernels, one of them is a shared convolution kernel. The
visible and infrared images are respectively convolved with different
spliced convolution kernels. The feature extraction layer of IVFuseNet
can be expressed as:

\begin{equation}
\begin{array}{c}
\sigma\left(\left(\left[\begin{array}{ccc}
W_{RGB} & W_{share} & W_{T}\end{array}\right]\left[\begin{array}{cccc}
1\\
 & 1 & 1\\
 &  &  & 1
\end{array}\right]\right)\ast\left[\begin{array}{cc}
F_{RGB}\\
 & F_{T}
\end{array}\right]\right)\\
=\left[\begin{array}{cc}
\sigma\left(\left[\begin{array}{cc}
W_{RGB} & W_{share}\end{array}\right]\ast F_{RGB}\right) & \sigma\left(\left[\begin{array}{cc}
W_{share} & W_{T}\end{array}\right]\ast F_{T}\right)\end{array}\right]
\end{array}
\end{equation}
Since the shared and non-shared convolution kernels are fused in convolution
kernel space, IVFuseNet only needs to perform two convolution operations.
However, due to the shared and non-shared convolution kernels are
concatenated in advance, the size of the convolution kernel is smaller
than that of MANet, which means IVFuseNet has weak representation
power than MANet. For example, in MANet, the size of the shared and
non-shared convolution kernel in the first layer is $96\times3\times7\times7$
and $96\times3\times3\times3$; while in IVFuseNet, the size of the
shared and non-shared convolution kernel in the corresponding layer
is $24\times3\times7\times7$ and $72\times3\times3\times3$. Besides,
the channel size of the shared and non-shared convolution kernel needs
to be prefixed, and the coupling rate cannot be adjusted in real-time
in the face of different contributions of individual features and
common features.

We summarize the related works below:
\begin{enumerate}
\item the speed of ConvNet is fast, but ConvNet does not have shared convolution
kernel to ectract common features.
\item Although MANet has both shared and non-shared convolution kernels,
the speed of MANet is much slower than that of ConvNet. Moreover,
MANet has no design to deal with the different contributions of the
individual features and common features.
\item IVFuseNet have both shared and non-shared convolution kernels, and
the speed of IVFuseNet is fast. However, compared with MANet, IVFuseNet
has weak representation power. Moreover, IVFuseNet also has no procedure
to handle the different contributions of the individual features and
common features.
\end{enumerate}

\section{Our Method}

In this section, we will introduce a novel RGBT tracking method called
dynamic fusion network (DFNet). We first introduce the dynamic fusion
layer. Dynamic convolution is used in the convolution kernel space
to fuse shared and non-shared convolution kernels. Then we use the
dynamic fusion layer as the basic module to construct DFNet for RGBT
tracking. DFNet has the advantages of MANet and IVFuseNet, which has
shared convolution kernels to extract common features. We highlight
the differences of the network structures between DFNet and the related
models in Table \ref{tab:1}. Due to the fusion of shared and non-shared
convolution kernels in convolution kernel space, DFNet has high inference
efficiency. Besides, adaptive convolutional features can be extracted
in the face of changes in the scene because of its dynamic nature.

\subsection{Dynamic Fusion Layer}

Dynamic fusion layer fuses the shared convolution kernel and non-shared
convolution kernels in convolution kernel space. That is, the convolution
kernels are merged, then the convolution operation is performed:
\begin{equation}
\begin{array}{c}
\sigma\left(\left(\left[\begin{array}{ccc}
W_{RGB} & W_{share} & W_{T}\end{array}\right]\left[\begin{array}{cc}
a\\
b & c\\
 & d
\end{array}\right]\right)\ast\left[\begin{array}{cc}
F_{RGB}\\
 & F_{T}
\end{array}\right]\right)\\
=\left[\begin{array}{cc}
\sigma\left(\left(aW_{RGB}+bW_{share}\right)\ast F_{RGB}\right) & \sigma\left(\left(cW_{share}+dW_{T}\right)\ast F_{T}\right)\end{array}\right]
\end{array}
\end{equation}
Its structure diagram is shown in Figure \ref{fig:2} (d). In feature
extraction, dynamic fusion layer only needs to perform two convolution
operations on the visible and infrared inputs respectively to obtain
common features and individual features, which greatly reduces computational
cost.

The fusion of shared and non-shared convolution kernels is a weighted
addition:
\begin{equation}
\begin{array}{c}
\left\{ \begin{array}{c}
\widetilde{W}_{RGB}=aW_{RGB}+bW_{share}\\
\widetilde{W}_{T}=cW_{share}+dW_{T}
\end{array}\right.\\
\textrm{s.t.}\:0<\left\{ a,b,c,d\right\} <1\:a+b=1\:c+d=1
\end{array}
\end{equation}
In this way, the size of the convolution kernels is not changed, and
no additional artificially set parameters are introduced. The weights
$a$, $b$, $c$, and $d$ are adaptive, which can be obtained from
the input through Global Average Pooling (GAP), two-layer full connection
(FC), and softmax:
\begin{equation}
\begin{array}{c}
\left\{ \begin{array}{c}
\left[a,b\right]=\mathcal{F}\left(F_{RGB}\right)\\
\left[c,d\right]=\mathcal{F}\left(F_{T}\right)
\end{array}\right.\\
\mathcal{F}\left(X\right)=\textrm{Softmax}\circ\textrm{FC}\circ\textrm{ReLU}\circ\textrm{GAP}(X)
\end{array}
\end{equation}
The weights are input-dependent so that the dynamic fusion layer can
use different convolution kernels for different image pairs, which
enhances the expressive ability of the model. Compared with the convolution
layer without weights, the dynamic fusion layer only increases the
Mult-Adds of 0.02\%, which can guarantee the real-time performance
of the network. For details, please refer to Section \ref{subsec:Efficiency-Analysis}. 

In the dynamic fusion layer, the convolution kernel is updated through
the back-propagation algorithm. In each iteration, the convolution
kernel of visible and infrared features iterates as follows: 
\begin{equation}
\begin{array}{c}
\widetilde{W}_{RGB}^{(i)}=a\left(W_{RGB}^{(i-1)}+\textrm{lr}\frac{\partial L}{\partial W_{RGB}^{(i-1)}}\right)+b\left(W_{share}^{(2i-1)}+\textrm{lr}\frac{\partial L}{\partial W_{share}^{(2i-1)}}\right)\end{array}
\end{equation}

\begin{equation}
\widetilde{W}_{T}^{(i)}=c\left(W_{share}^{(2i-2)}+\textrm{lr}\frac{\partial L}{\partial W_{share}^{(2i-2)}}\right)+d\left(W_{T}^{(i-1)}+\textrm{lr}\frac{\partial L}{\partial W_{T}^{(i-1)}}\right)
\end{equation}
where $i$ is the iteration numbers, $\textrm{lr}$ is the learning
rate, and $L$ is the loss function. In each iteration, the non-shared
convolution kernels are updated once, and the shared convolution kernel
are updated twice.

\subsection{The Architecture}

\begin{figure}
\begin{centering}
\includegraphics[scale=0.4]{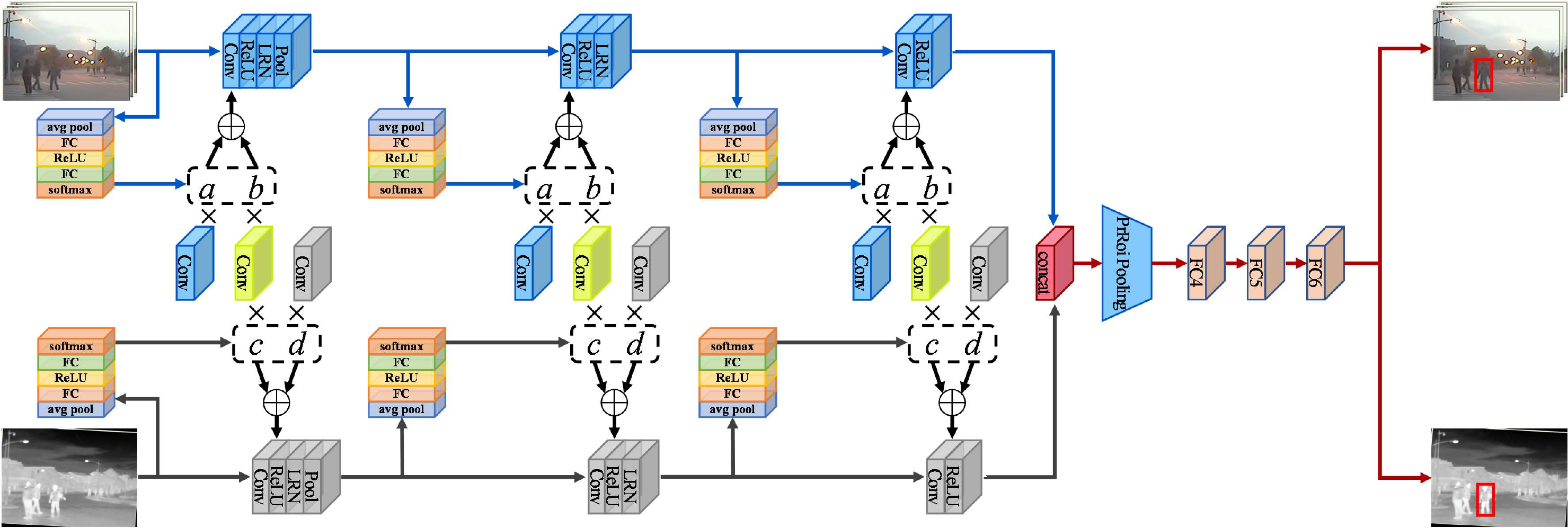}
\par\end{centering}
\caption{\label{fig:3}The overall architecture of DFNet. DFNet uses a multi-domain
learning framework. Three dynamic fusion layers are used to extract
and fuse the features of visible and infrared images. PrRoiPooling
is used to unify the features into a $3\times3$ size. Three fully
connected layers are used to determine whether the candidate is the
object or background.}
\end{figure}

The overall architecture of DFNet is shown in Figure \ref{fig:3}.
The features of visible and infrared images are first extracted and
fused through three dynamic fusion layers. After PrRoiPooling \cite{ref20},
the features of different sizes are unified into $3\times3$. Then,
features enter three fully connected networks to determine the object
or background. At the end of the tracking process, DFNet takes the
candidate with the highest network output score as the object:
\begin{equation}
x_{t}^{\ast}=\textrm{arg}\,\max\mathscr{F}\left(x_{t}^{i}\right)
\end{equation}
where $x_{t}^{i}$ represents the $i$-th candidate frame in the $t$-th
frame, $\mathscr{F}\left(\cdot\right)$ represents the score of network
output, and $x_{t}^{*}$ represents the final object result of the
$t$-th frame.

Inspired by RT-MDNet \cite{ref21}, DFNet adopts a multi-domain learning
framework. During training, all video sequences share three dynamic
fusion layers, FC4 and FC5. Each video sequence uses a domain-specific
FC6. During testing, the multiple domain-specific FC6s are replaced
with a reinitialized FC6.

\section{Implementation Details}

We train and test DFNet on the PyTorch platform with i7-10700K CPU
and TITAN RTX GPU. We will introduce the details of training and online
tracking process in this section.

\subsection{Offline Training}

The pre-trained network in VGG-M \cite{ref22} is adopted to initialize
the model and use ImageNet \cite{ref29} and RGBT (GTOT \cite{ref23}
or RGBT234 \cite{ref24}) mixed dataset to train DFNet. We train the
network using stochastic gradient descent with momentum. The momentum
is set to 0.9, the learning rate is set to 1e-4, and the weight decay
is set to 5e-4. The number of epochs is set to 60.

\subsection{Online Tracking}

In the online tracking phase, we initialize the model with the trained
three dynamic fusion layers, FC4, and FC5. We reinitialize a new FC6
and use the first frame to train FC6. Specifically, we collect 500
positive samples (IOU>0.7) and 5000 negative samples (IOU<0.3) from
the first frame as training samples, and use stochastic gradient descent
with momentum for training. The momentum is set to 0.9, the learning
rate is set to 1e-4, and the weight decay is set to 5e-4.

In the follow-up tracking phase, three dynamic fusion layers are fixed,
but FC4, FC5, and FC6 are fine-tuned online. We collect 50 positive
samples and 200 negative samples, perform long-term updates every
10 frames, and perform short-term updates when tracking fails. The
learning rate of FC6 is set to 1e-3, and the learning rate of FC4
and FC5 is set to 5e-4. At time $t$, we use a Gaussian sampler to
collect 256 candidates around the object position in the previous
frame and calculate their respective classification scores through
the network. Then, the Multi-layer Perceptron (MLP) is used to regress
the average value of the top five bounding boxes of the classification
score to obtain the final bounding box.

\section{Experiment}

\subsection{Dataset and Evaluation Matrix}

We use two RGBT datasets, GTOT \cite{ref23} and RGBT234, \cite{ref24}
to compare DFNet with other tracking methods. We use ImageNet and
GTOT mixed dataset as the training set when evaluating on RGBT234;
we use ImageNet and RGBT234 mixed dataset as the training set when
evaluating on GTOT. The GTOT and RGBT234 datasets have 50 and 234
RGBT sequences of image pairs aligned in space and time, respectively.
In one-pass evaluation (OPE), we use Precision Rate (PR) and Success
Rate (SR) as evaluation indicators to evaluate the tracking results.
PR refers to the proportion of frames whose difference between the
output position and the ground truth bounding box is within the threshold.
The thresholds of GTOT is set to 5, The thresholds of RGBT234 is set
to 20. SR is the proportion of frames where the overlap ratio between
the output position and the ground truth bounding box is greater than
the threshold. The area under the curves (AUC) is employed to calculate
the SR score.

\subsection{Comparison with Other Methods}

\begin{figure}
\subfloat[comparison on GTOT]{\includegraphics[scale=0.27]{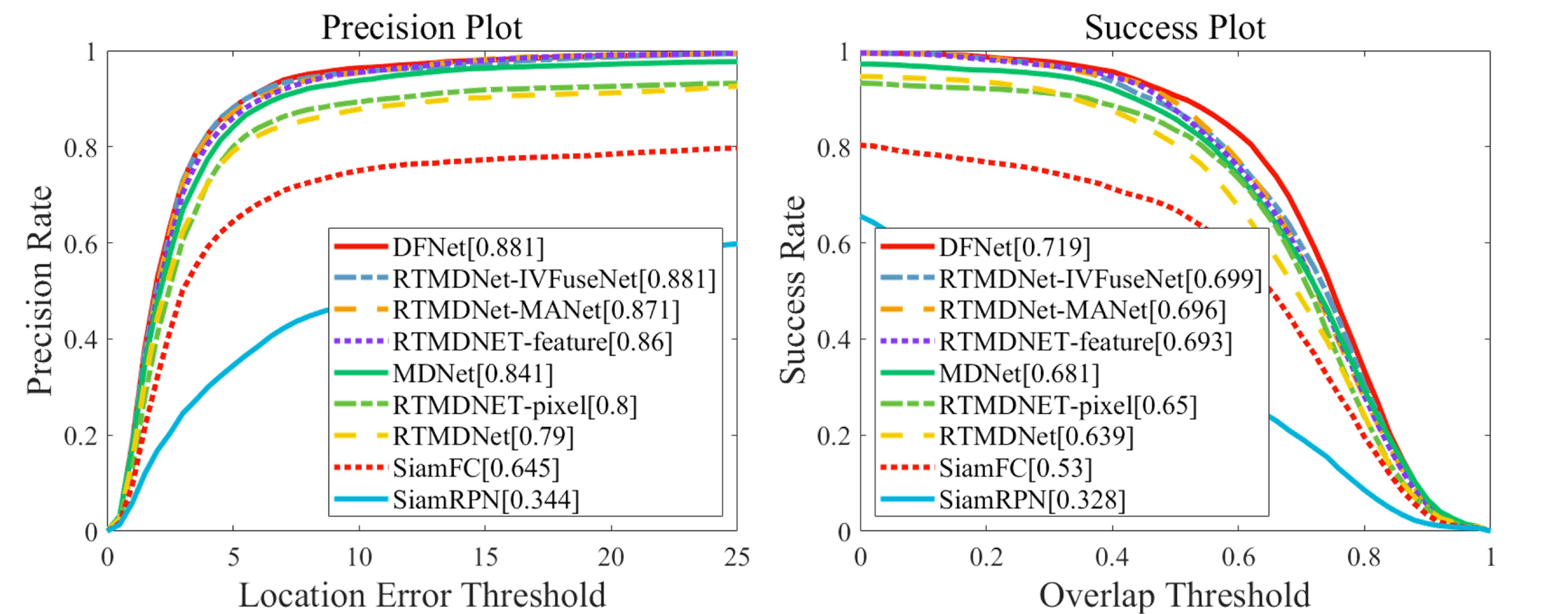}

}

\subfloat[comparison on RGBT234]{\includegraphics[scale=0.27]{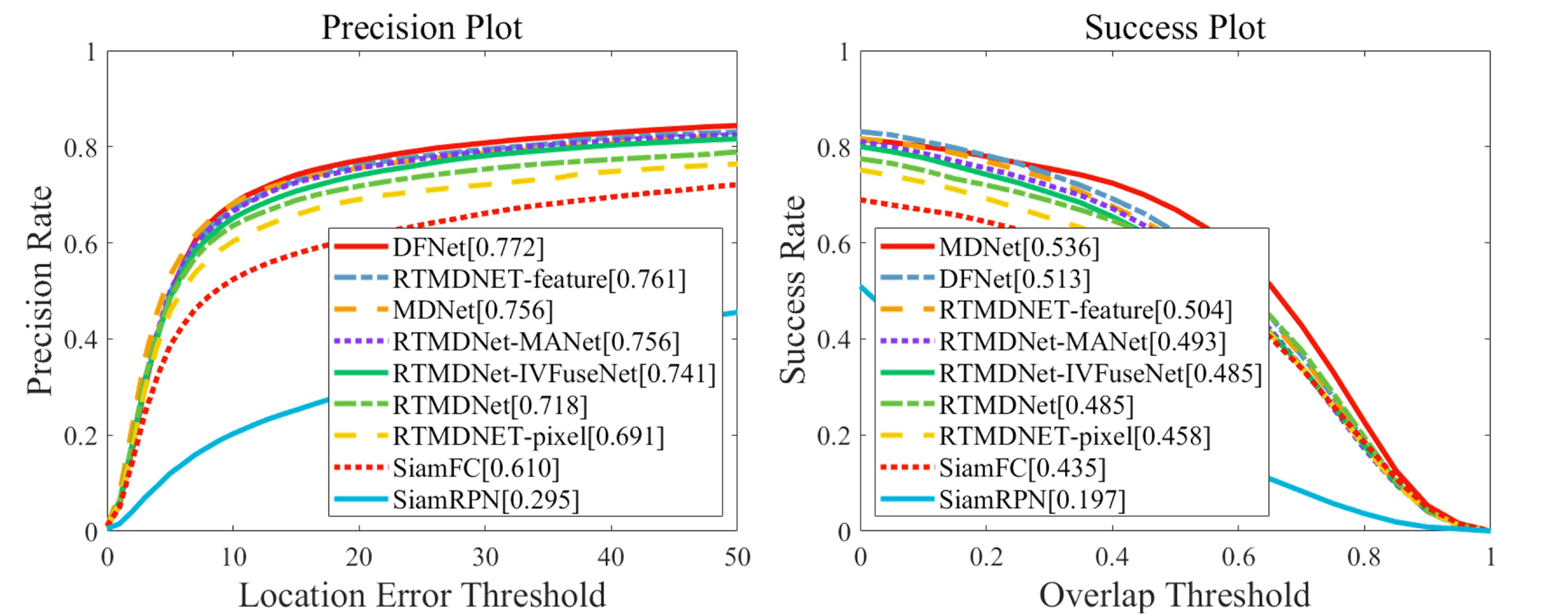}

}

\caption{\label{fig:4}Overall performance compared with other trackers on
GTOT (a) and RGBT234 (b). }

\end{figure}

We compared DFNet with visible tracking methods (MDNet \cite{ref25},
RT-MDNet \cite{ref21}, SiamFC \cite{ref26}, and SiamRPN \cite{ref27})
and fusion tracking methods (pixel-level fusion \cite{pixel1,pixel2,ref28},
feature-level fusion \cite{ref9,ref11}, MANet \cite{ref15}, and
IVFuseNet \cite{ref16}). To be fair, all fusion methods have been
implemented on the RT-MDNet tracking framework, represented below
as RTMDNet-pixel, RTMDNet-feature, RTMDNet-MANet, and RTMDNet-IVFuseNet,
respectively. The overall tracking performance is shown in Figure
\ref{fig:4}. For all the indicators of these two benchmarks, our
DFNet has clearly outperformed other tracking methods. Specifically,
on the GTOT benchmark, our DFNet reached 88.1\%/71.9\% on PR/SR. While
on the RGBT234 benchmark, our DFNet reached PR/SR 77.2\%/51.3\%. To
further show the effectiveness of DFNet, we list the performance of
each attribute of the RGBT234 dataset. The specific tracking results
are shown in Table \ref{tab:2}. It can be concluded from the table
that our proposed DFNet outperforms other trackers in 8 cases with
higher PR. 

\begin{table}
\caption{\label{tab:2}RGBT234 dataset PR/SR scores based on attributes. The
best, second-best, and third-best PR/SR are shown in \textcolor{red}{red},
\textcolor{orange}{yellow}, and \textcolor{blue}{blue}.}

\selectlanguage{british}%
\centering{}\resizebox{\textwidth}{!}{
\begin{tabular}{c|cccccccc|c}
\hline 
\selectlanguage{english}%
{\scriptsize{}Tracker}\selectlanguage{british}%
 & \selectlanguage{english}%
{\scriptsize{}SiamFC}\selectlanguage{british}%
 & \selectlanguage{english}%
{\scriptsize{}SiamRPN}\selectlanguage{british}%
 & \selectlanguage{english}%
{\scriptsize{}MDNet}\selectlanguage{british}%
 & \selectlanguage{english}%
{\scriptsize{}RTMDNet}\selectlanguage{british}%
 & \selectlanguage{english}%
{\scriptsize{}}%
\begin{tabular}{@{}c@{}}
{\scriptsize{}RTMDNet}\tabularnewline
{\scriptsize{}-pixel}\tabularnewline
\end{tabular}\selectlanguage{british}%
 & \selectlanguage{english}%
{\scriptsize{}}%
\begin{tabular}{@{}c@{}}
{\scriptsize{}RTMDNet}\tabularnewline
{\scriptsize{}-feature}\tabularnewline
\end{tabular}\selectlanguage{british}%
 & \selectlanguage{english}%
{\scriptsize{}}%
\begin{tabular}{@{}c@{}}
{\scriptsize{}RTMDNet}\tabularnewline
{\scriptsize{}-IVFuseNet}\tabularnewline
\end{tabular}\selectlanguage{british}%
 & \selectlanguage{english}%
{\scriptsize{}}%
\begin{tabular}{@{}c@{}}
{\scriptsize{}RTMDNet}\tabularnewline
{\scriptsize{}-MANet}\tabularnewline
\end{tabular}\selectlanguage{british}%
 & \selectlanguage{english}%
{\scriptsize{}DFNet}\selectlanguage{british}%
\tabularnewline
\hline 
\selectlanguage{english}%
{\scriptsize{}BC}\selectlanguage{british}%
 & \selectlanguage{english}%
{\scriptsize{}0.496/0.333}\selectlanguage{british}%
 & \selectlanguage{english}%
{\scriptsize{}0.187/0.116}\selectlanguage{british}%
 & \selectlanguage{english}%
{\scriptsize{}0.683/}\textcolor{red}{\scriptsize{}0.462}\selectlanguage{british}%
 & \selectlanguage{english}%
{\scriptsize{}0.630/0.402}\selectlanguage{british}%
 & \selectlanguage{english}%
{\scriptsize{}0.582/0.375}\selectlanguage{british}%
 & \selectlanguage{english}%
\textcolor{orange}{\scriptsize{}0.705}{\scriptsize{}/}\textcolor{blue}{\scriptsize{}0.439}\selectlanguage{british}%
 & \selectlanguage{english}%
{\scriptsize{}0.659/0.397}\selectlanguage{british}%
 & \selectlanguage{english}%
\textcolor{blue}{\scriptsize{}0.697}{\scriptsize{}/0.437}\selectlanguage{british}%
 & \selectlanguage{english}%
\textcolor{red}{\scriptsize{}0.714}{\scriptsize{}/}\textcolor{orange}{\scriptsize{}0.452}\selectlanguage{british}%
\tabularnewline
\selectlanguage{english}%
{\scriptsize{}CM}\selectlanguage{british}%
 & \selectlanguage{english}%
{\scriptsize{}0.564/0.407}\selectlanguage{british}%
 & \selectlanguage{english}%
{\scriptsize{}0.321/0.226}\selectlanguage{british}%
 & \selectlanguage{english}%
\textcolor{blue}{\scriptsize{}0.689}{\scriptsize{}/}\textcolor{red}{\scriptsize{}0.497}\selectlanguage{british}%
 & \selectlanguage{english}%
{\scriptsize{}0.637/0.438}\selectlanguage{british}%
 & \selectlanguage{english}%
{\scriptsize{}0.626/0.429}\selectlanguage{british}%
 & \selectlanguage{english}%
\textcolor{orange}{\scriptsize{}0.690}{\scriptsize{}/}\textcolor{blue}{\scriptsize{}0.467}\selectlanguage{british}%
 & \selectlanguage{english}%
{\scriptsize{}0.663/0.438}\selectlanguage{british}%
 & \selectlanguage{english}%
{\scriptsize{}0.676/0.448}\selectlanguage{british}%
 & \selectlanguage{english}%
\textcolor{red}{\scriptsize{}0.692}{\scriptsize{}/}\textcolor{orange}{\scriptsize{}0.471}\selectlanguage{british}%
\tabularnewline
\selectlanguage{english}%
{\scriptsize{}DEF}\selectlanguage{british}%
 & \selectlanguage{english}%
{\scriptsize{}0.591/0.431}\selectlanguage{british}%
 & \selectlanguage{english}%
{\scriptsize{}0.281/0.212}\selectlanguage{british}%
 & \selectlanguage{english}%
\textcolor{red}{\scriptsize{}0.685}{\scriptsize{}/}\textcolor{red}{\scriptsize{}0.497}\selectlanguage{british}%
 & \selectlanguage{english}%
{\scriptsize{}0.654/0.451}\selectlanguage{british}%
 & \selectlanguage{english}%
{\scriptsize{}0.611/0.419}\selectlanguage{british}%
 & \selectlanguage{english}%
\textcolor{blue}{\scriptsize{}0.679}{\scriptsize{}/}\textcolor{orange}{\scriptsize{}0.466}\selectlanguage{british}%
 & \selectlanguage{english}%
{\scriptsize{}0.651/0.448}\selectlanguage{british}%
 & \selectlanguage{english}%
\textcolor{orange}{\scriptsize{}0.682}{\scriptsize{}/0.445}\selectlanguage{british}%
 & \selectlanguage{english}%
{\scriptsize{}0.661/}\textcolor{blue}{\scriptsize{}0.462}\selectlanguage{british}%
\tabularnewline
\selectlanguage{english}%
{\scriptsize{}FM}\selectlanguage{british}%
 & \selectlanguage{english}%
{\scriptsize{}0.518/0.374}\selectlanguage{british}%
 & \selectlanguage{english}%
{\scriptsize{}0.276/0.155}\selectlanguage{british}%
 & \selectlanguage{english}%
\textcolor{red}{\scriptsize{}0.690}{\scriptsize{}/}\textcolor{red}{\scriptsize{}0.448}\selectlanguage{british}%
 & \selectlanguage{english}%
\textcolor{orange}{\scriptsize{}0.679}{\scriptsize{}/}\textcolor{orange}{\scriptsize{}0.404}\selectlanguage{british}%
 & \selectlanguage{english}%
{\scriptsize{}0.595/0.330}\selectlanguage{british}%
 & \selectlanguage{english}%
{\scriptsize{}0.637/0.365}\selectlanguage{british}%
 & \selectlanguage{english}%
{\scriptsize{}0.618/0.356}\selectlanguage{british}%
 & \selectlanguage{english}%
{\scriptsize{}0.621/0.374}\selectlanguage{british}%
 & \selectlanguage{english}%
\textcolor{blue}{\scriptsize{}0.640}{\scriptsize{}/}\textcolor{blue}{\scriptsize{}0.378}\selectlanguage{british}%
\tabularnewline
\selectlanguage{english}%
{\scriptsize{}HO}\selectlanguage{british}%
 & \selectlanguage{english}%
{\scriptsize{}0.521/0.367}\selectlanguage{british}%
 & \selectlanguage{english}%
{\scriptsize{}0.261/0.164}\selectlanguage{british}%
 & \selectlanguage{english}%
\textcolor{red}{\scriptsize{}0.654}{\scriptsize{}/}\textcolor{red}{\scriptsize{}0.459}\selectlanguage{british}%
 & \selectlanguage{english}%
{\scriptsize{}0.634/}\textcolor{orange}{\scriptsize{}0.422}\selectlanguage{british}%
 & \selectlanguage{english}%
{\scriptsize{}0.586/0.369}\selectlanguage{british}%
 & \selectlanguage{english}%
{\scriptsize{}0.621/0.403}\selectlanguage{british}%
 & \selectlanguage{english}%
{\scriptsize{}0.592/0.381}\selectlanguage{british}%
 & \selectlanguage{english}%
\textcolor{orange}{\scriptsize{}0.644}{\scriptsize{}/0.409}\selectlanguage{british}%
 & \selectlanguage{english}%
\textcolor{blue}{\scriptsize{}0.641}{\scriptsize{}/}\textcolor{blue}{\scriptsize{}0.412}\selectlanguage{british}%
\tabularnewline
\selectlanguage{english}%
{\scriptsize{}LI}\selectlanguage{british}%
 & \selectlanguage{english}%
{\scriptsize{}0.495/0.356}\selectlanguage{british}%
 & \selectlanguage{english}%
{\scriptsize{}0.231/0.154}\selectlanguage{british}%
 & \selectlanguage{english}%
{\scriptsize{}0.674/0.451}\selectlanguage{british}%
 & \selectlanguage{english}%
{\scriptsize{}0.605/0.391}\selectlanguage{british}%
 & \selectlanguage{english}%
{\scriptsize{}0.609/0.404}\selectlanguage{british}%
 & \selectlanguage{english}%
\textcolor{orange}{\scriptsize{}0.763}{\scriptsize{}/}\textcolor{orange}{\scriptsize{}0.504}\selectlanguage{british}%
 & \selectlanguage{english}%
{\scriptsize{}0.742/0.492}\selectlanguage{british}%
 & \selectlanguage{english}%
\textcolor{blue}{\scriptsize{}0.756}{\scriptsize{}/}\textcolor{blue}{\scriptsize{}0.497}\selectlanguage{british}%
 & \selectlanguage{english}%
\textcolor{red}{\scriptsize{}0.789}{\scriptsize{}/}\textcolor{red}{\scriptsize{}0.528}\selectlanguage{british}%
\tabularnewline
\selectlanguage{english}%
{\scriptsize{}LR}\selectlanguage{british}%
 & \selectlanguage{english}%
{\scriptsize{}0.603/0.404}\selectlanguage{british}%
 & \selectlanguage{english}%
{\scriptsize{}0.295/0.159}\selectlanguage{british}%
 & \selectlanguage{english}%
{\scriptsize{}0.734/}\textcolor{orange}{\scriptsize{}0.502}\selectlanguage{british}%
 & \selectlanguage{english}%
{\scriptsize{}0.683/0.447}\selectlanguage{british}%
 & \selectlanguage{english}%
{\scriptsize{}0.727/0.464}\selectlanguage{british}%
 & \selectlanguage{english}%
\textcolor{orange}{\scriptsize{}0.794}{\scriptsize{}/}\textcolor{blue}{\scriptsize{}0.492}\selectlanguage{british}%
 & \selectlanguage{english}%
{\scriptsize{}0.787/0.471}\selectlanguage{british}%
 & \selectlanguage{english}%
{\scriptsize{}0.730/0.446}\selectlanguage{british}%
 & \selectlanguage{english}%
\textcolor{red}{\scriptsize{}0.797}{\scriptsize{}/}\textcolor{orange}{\scriptsize{}0.496}\selectlanguage{british}%
\tabularnewline
\selectlanguage{english}%
{\scriptsize{}MB}\selectlanguage{british}%
 & \selectlanguage{english}%
{\scriptsize{}0.554/0.405}\selectlanguage{british}%
 & \selectlanguage{english}%
{\scriptsize{}0.310/0.209}\selectlanguage{british}%
 & \selectlanguage{english}%
\textcolor{red}{\scriptsize{}0.702}{\scriptsize{}/}\textcolor{red}{\scriptsize{}0.517}\selectlanguage{british}%
 & \selectlanguage{english}%
{\scriptsize{}0.669/0.467}\selectlanguage{british}%
 & \selectlanguage{english}%
{\scriptsize{}0.633/0.442}\selectlanguage{british}%
 & \selectlanguage{english}%
\textcolor{blue}{\scriptsize{}0.676}{\scriptsize{}/}\textcolor{blue}{\scriptsize{}0.470}\selectlanguage{british}%
 & \selectlanguage{english}%
{\scriptsize{}0.670/0.450}\selectlanguage{british}%
 & \selectlanguage{english}%
{\scriptsize{}0.635/0.433}\selectlanguage{british}%
 & \selectlanguage{english}%
\textcolor{red}{\scriptsize{}0.702}{\scriptsize{}/}\textcolor{orange}{\scriptsize{}0.489}\selectlanguage{british}%
\tabularnewline
\selectlanguage{english}%
{\scriptsize{}NO}\selectlanguage{british}%
 & \selectlanguage{english}%
{\scriptsize{}0.765/0.564}\selectlanguage{british}%
 & \selectlanguage{english}%
{\scriptsize{}0.404/0.282}\selectlanguage{british}%
 & \selectlanguage{english}%
\textcolor{blue}{\scriptsize{}0.862}{\scriptsize{}/}\textcolor{red}{\scriptsize{}0.636}\selectlanguage{british}%
 & \selectlanguage{english}%
{\scriptsize{}0.842/0.576}\selectlanguage{british}%
 & \selectlanguage{english}%
{\scriptsize{}0.828/0.557}\selectlanguage{british}%
 & \selectlanguage{english}%
{\scriptsize{}0.859/}\textcolor{blue}{\scriptsize{}0.582}\selectlanguage{british}%
 & \selectlanguage{english}%
{\scriptsize{}0.856/0.564}\selectlanguage{british}%
 & \selectlanguage{english}%
\textcolor{orange}{\scriptsize{}0.868}{\scriptsize{}/0.569}\selectlanguage{british}%
 & \selectlanguage{english}%
\textcolor{red}{\scriptsize{}0.871}{\scriptsize{}/}\textcolor{orange}{\scriptsize{}0.599}\selectlanguage{british}%
\tabularnewline
\selectlanguage{english}%
{\scriptsize{}PO}\selectlanguage{british}%
 & \selectlanguage{english}%
{\scriptsize{}0.629/0.446}\selectlanguage{british}%
 & \selectlanguage{english}%
{\scriptsize{}0.275/0.188}\selectlanguage{british}%
 & \selectlanguage{english}%
{\scriptsize{}0.810/}\textcolor{blue}{\scriptsize{}0.567}\selectlanguage{british}%
 & \selectlanguage{english}%
{\scriptsize{}0.754/0.513}\selectlanguage{british}%
 & \selectlanguage{english}%
{\scriptsize{}0.714/0.508}\selectlanguage{british}%
 & \selectlanguage{english}%
\textcolor{orange}{\scriptsize{}0.856}{\scriptsize{}/}\textcolor{orange}{\scriptsize{}0.569}\selectlanguage{british}%
 & \selectlanguage{english}%
{\scriptsize{}0.838/0.554}\selectlanguage{british}%
 & \selectlanguage{english}%
{\scriptsize{}0.817/0.544}\selectlanguage{british}%
 & \selectlanguage{english}%
\textcolor{red}{\scriptsize{}0.857}{\scriptsize{}/}\textcolor{red}{\scriptsize{}0.575}\selectlanguage{british}%
\tabularnewline
\selectlanguage{english}%
{\scriptsize{}SV}\selectlanguage{british}%
 & \selectlanguage{english}%
{\scriptsize{}0.634/0.461}\selectlanguage{british}%
 & \selectlanguage{english}%
{\scriptsize{}0.308/0.210}\selectlanguage{british}%
 & \selectlanguage{english}%
\textcolor{red}{\scriptsize{}0.767}{\scriptsize{}/}\textcolor{red}{\scriptsize{}0.549}\selectlanguage{british}%
 & \selectlanguage{english}%
{\scriptsize{}0.747/}\textcolor{orange}{\scriptsize{}0.508}\selectlanguage{british}%
 & \selectlanguage{english}%
{\scriptsize{}0.697/0.461}\selectlanguage{british}%
 & \selectlanguage{english}%
\textcolor{blue}{\scriptsize{}0.751}{\scriptsize{}/0.499}\selectlanguage{british}%
 & \selectlanguage{english}%
{\scriptsize{}0.743/0.494}\selectlanguage{british}%
 & \selectlanguage{english}%
\textcolor{orange}{\scriptsize{}0.762}{\scriptsize{}/}\textcolor{blue}{\scriptsize{}0.505}\selectlanguage{british}%
 & \selectlanguage{english}%
{\scriptsize{}0.749/0.501}\selectlanguage{british}%
\tabularnewline
\selectlanguage{english}%
{\scriptsize{}TC}\selectlanguage{british}%
 & \selectlanguage{english}%
{\scriptsize{}0.681/0.488}\selectlanguage{british}%
 & \selectlanguage{english}%
{\scriptsize{}0.233/0.155}\selectlanguage{british}%
 & \selectlanguage{english}%
\textcolor{red}{\scriptsize{}0.801}{\scriptsize{}/}\textcolor{red}{\scriptsize{}0.585}\selectlanguage{british}%
 & \selectlanguage{english}%
{\scriptsize{}0.763/}\textcolor{orange}{\scriptsize{}0.551}\selectlanguage{british}%
 & \selectlanguage{english}%
{\scriptsize{}0.727/0.487}\selectlanguage{british}%
 & \selectlanguage{english}%
\textcolor{blue}{\scriptsize{}0.771}{\scriptsize{}/0.520}\selectlanguage{british}%
 & \selectlanguage{english}%
{\scriptsize{}0.737/0.490}\selectlanguage{british}%
 & \selectlanguage{english}%
{\scriptsize{}0.743/0.490}\selectlanguage{british}%
 & \selectlanguage{english}%
\textcolor{orange}{\scriptsize{}0.796}{\scriptsize{}/}\textcolor{blue}{\scriptsize{}0.543}\selectlanguage{british}%
\tabularnewline
\hline 
\selectlanguage{english}%
{\scriptsize{}ALL}\selectlanguage{british}%
 & \selectlanguage{english}%
{\scriptsize{}0.610/0.435}\selectlanguage{british}%
 & \selectlanguage{english}%
{\scriptsize{}0.295/0.197}\selectlanguage{british}%
 & \selectlanguage{english}%
\textcolor{blue}{\scriptsize{}0.756}{\scriptsize{}/}\textcolor{red}{\scriptsize{}0.536}\selectlanguage{british}%
 & \selectlanguage{english}%
{\scriptsize{}0.718/0.485}\selectlanguage{british}%
 & \selectlanguage{english}%
{\scriptsize{}0.691/0.458}\selectlanguage{british}%
 & \selectlanguage{english}%
\textcolor{orange}{\scriptsize{}0.761}{\scriptsize{}/}\textcolor{blue}{\scriptsize{}0.504}\selectlanguage{british}%
 & \selectlanguage{english}%
{\scriptsize{}0.741/0.485}\selectlanguage{british}%
 & \selectlanguage{english}%
{\scriptsize{}0.756/0.493}\selectlanguage{british}%
 & \selectlanguage{english}%
\textcolor{red}{\scriptsize{}0.772}{\scriptsize{}/}\textcolor{orange}{\scriptsize{}0.513}\selectlanguage{british}%
\tabularnewline
\hline 
\end{tabular}}\selectlanguage{english}%
\end{table}

\subsection{Ablation Study}

\subsubsection{The Importance of Fusion}

\begin{figure}
\begin{centering}
\includegraphics[scale=0.27]{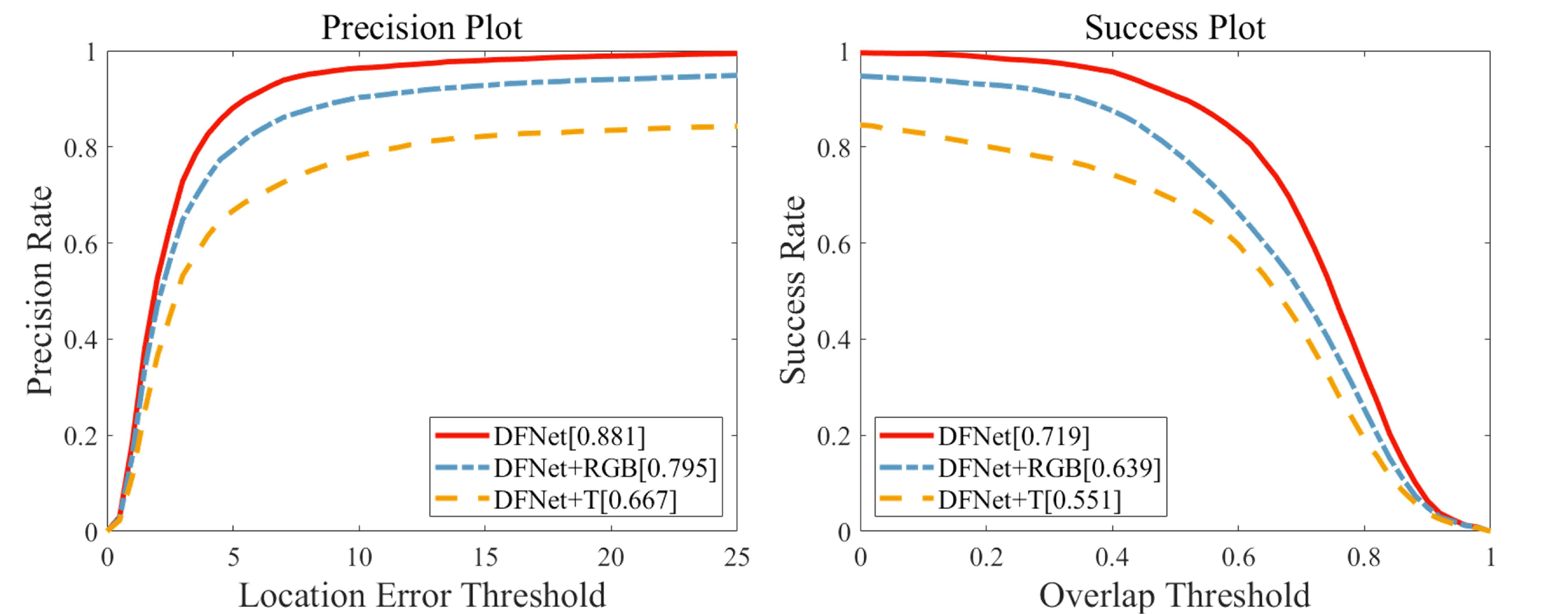}
\par\end{centering}
\caption{\label{fig:5}Comparison of visible, infrared, and fusion tracking}
\end{figure}

In order to show the importance of fusion for tracking, we compared
the tracking performance of DFNet+RGB, DFNet+T, and DFNet. DFNet+RGB
and DFNet+T respectively indicate that DFNet solely relies on visible
or infrared images for tracking. The tracking performance of DFNet
is shown in Figure \ref{fig:5}. The PR/SR of RGBT tracking are 8.6\%/8.0\%
higher than tracking using visible image alone, and 21.4\%/16.8\%
higher than tracking using infrared image alone. Experimental results
show that the performance of DFNet is significantly better than that
of methods based on single-modal images.

\subsubsection{Dynamic Fusion at Different Layers}

We perform a number of ablation studies on RT-MDNet to verify the
performance of the dynamic fusion layer at three different layers.
The results are shown in Table \ref{tab:3}. it can be found that
the more the dynamic fusion layer is used, the better the performance
is. Using dynamic fusion layers for all three layers produces the
best results. And the later the dynamic fusion layer used in the network,
the better the performance is.

\begin{table}
\caption{\label{tab:3}Dynamic fusion at different layers in RT-MDNet. \textsurd{}
indicates this layer uses a dynamic fusion layer to replace the vanilla
convolution, while - indicates not.}

\centering{}%
\begin{tabular}{cccccc}
\hline 
Network & C1 & C2 & C3 & PR & SR\tabularnewline
\hline 
Feature-level fusion & - & - & - & 0.860 & 0.693\tabularnewline
\hline 
 & \textsurd{} & - & - & 0.860 & 0.689\tabularnewline
Feature-level fusion & - & \textsurd{} & - & 0.863 & 0.695\tabularnewline
+ & - & - & \textsurd{} & 0.866 & 0.691\tabularnewline
Dynamic fusion layer & \textsurd{} & \textsurd{} & - & 0.867 & 0.699\tabularnewline
 & \textsurd{} & - & \textsurd{} & 0.866 & 0.699\tabularnewline
 & - & \textsurd{} & \textsurd{} & 0.872 & 0.702\tabularnewline
\hline 
DFNet & \textsurd{} & \textsurd{} & \textsurd{} & 0.881 & 0.709\tabularnewline
\hline 
\end{tabular}
\end{table}

\begin{figure}
\begin{centering}
\subfloat[Weights of visible images]{\begin{centering}
\includegraphics[scale=0.31]{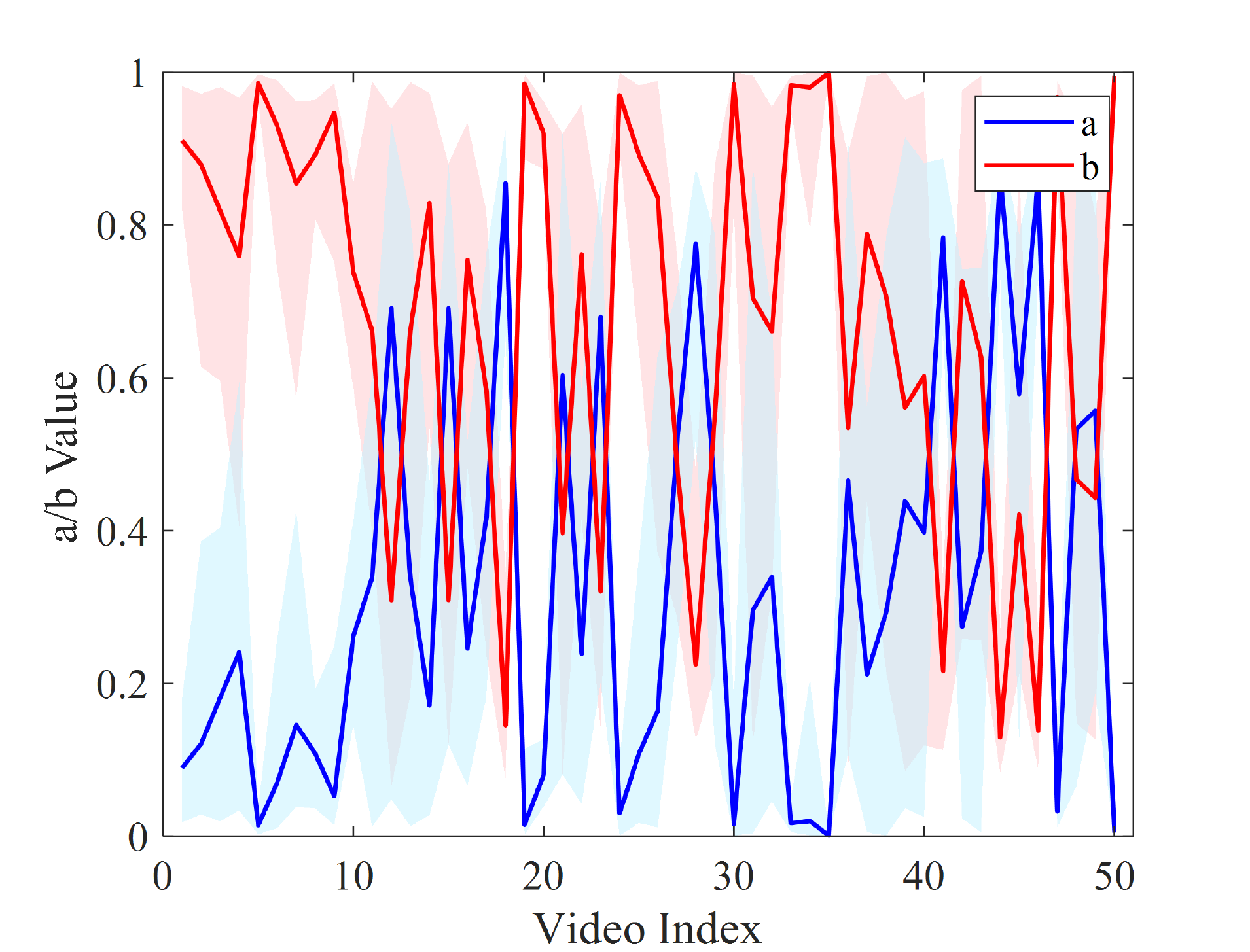}
\par\end{centering}
}\subfloat[Weights of infrared images]{\begin{centering}
\includegraphics[scale=0.31]{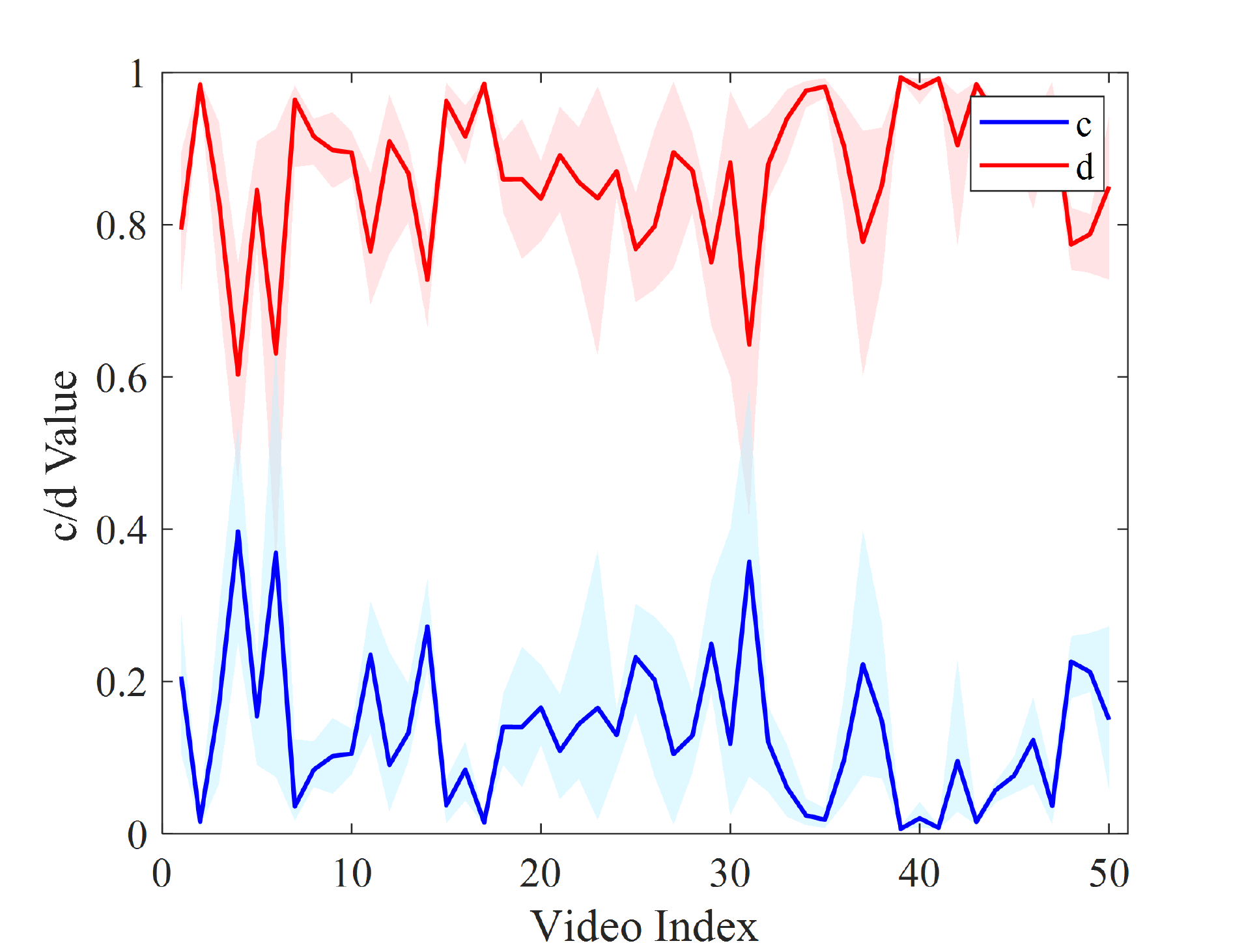}
\par\end{centering}
}
\par\end{centering}
\caption{\label{fig:6}Weights for different video sequences across the GTOT
dataset in DFNet. }
\end{figure}

We visualized the weights of the dynamic fusion layer in the order
of the videos in GTOT, as shown in Figure \ref{fig:6}. In Figure
\ref{fig:6} (a), the blue solid line represents the average of the
visible non-shared convolution kernel weights, and the blue shading
represents the range of the visible non-shared convolution kernel
weights. The red solid line represents the average of the visible
shared convolution kernel weights, and the red shade represents the
range of the visible shared convolution kernel weights. In Figure
\ref{fig:6} (b), the red solid line represents the average of the
infrared non-shared convolution kernel weights, and the red shade
represents the range of the infrared non-shared convolution kernel
weights. The blue solid line represents the average of the infrared
shared convolution kernel weights, and the blue shading represents
the range of the infrared shared convolution kernel weights. It can
be found that the dynamic fusion layer can calculate different weights
according to videos. In this way, the dynamic fusion layer makes the
fusion tracker adaptively calculates the contributions of individual
features and common features.

\begin{figure}
\begin{centering}
\subfloat[Weights of visible images]{\begin{centering}
\includegraphics[scale=0.4]{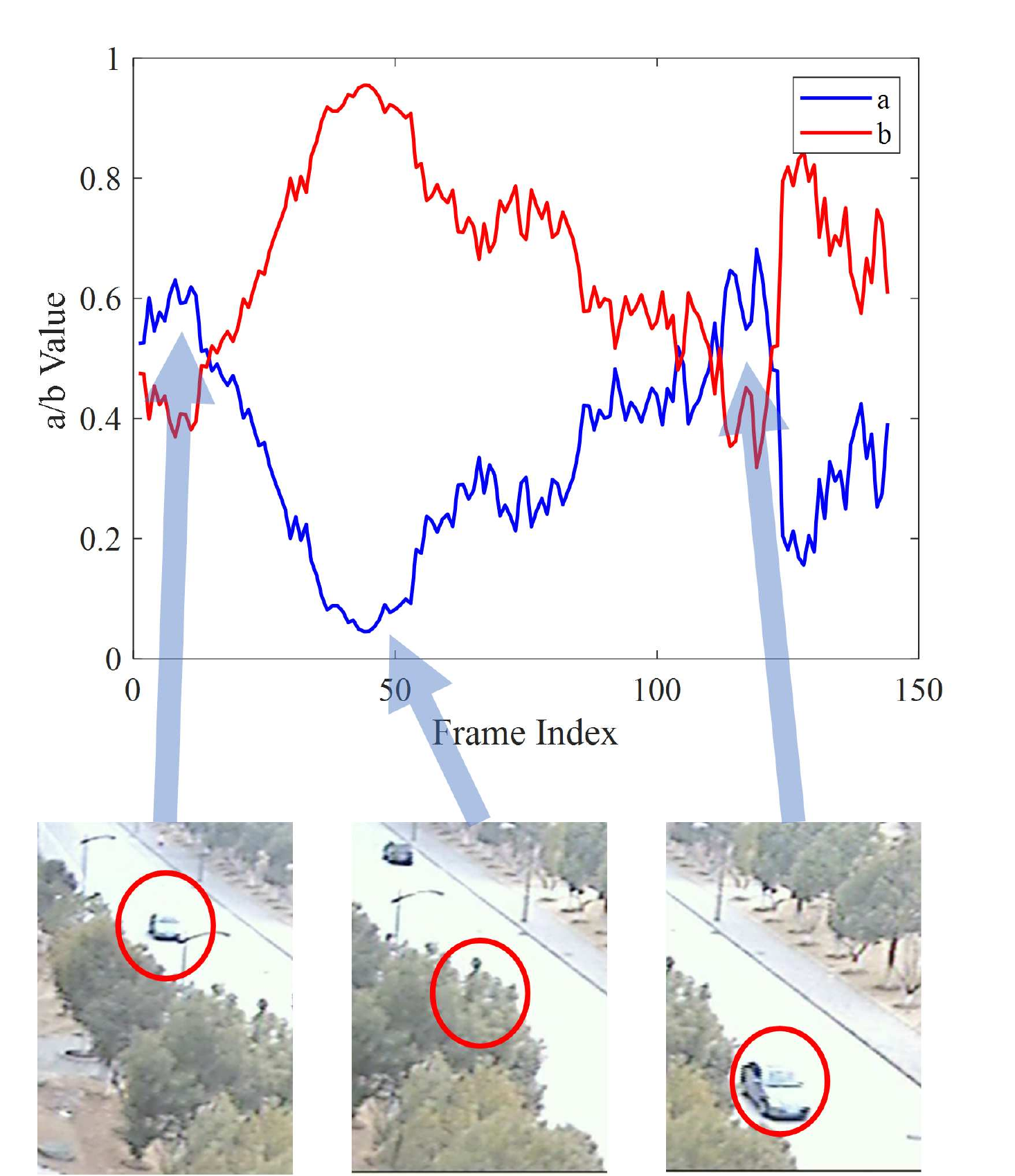}
\par\end{centering}
}\subfloat[Weights of infrared images]{\begin{centering}
\includegraphics[scale=0.4]{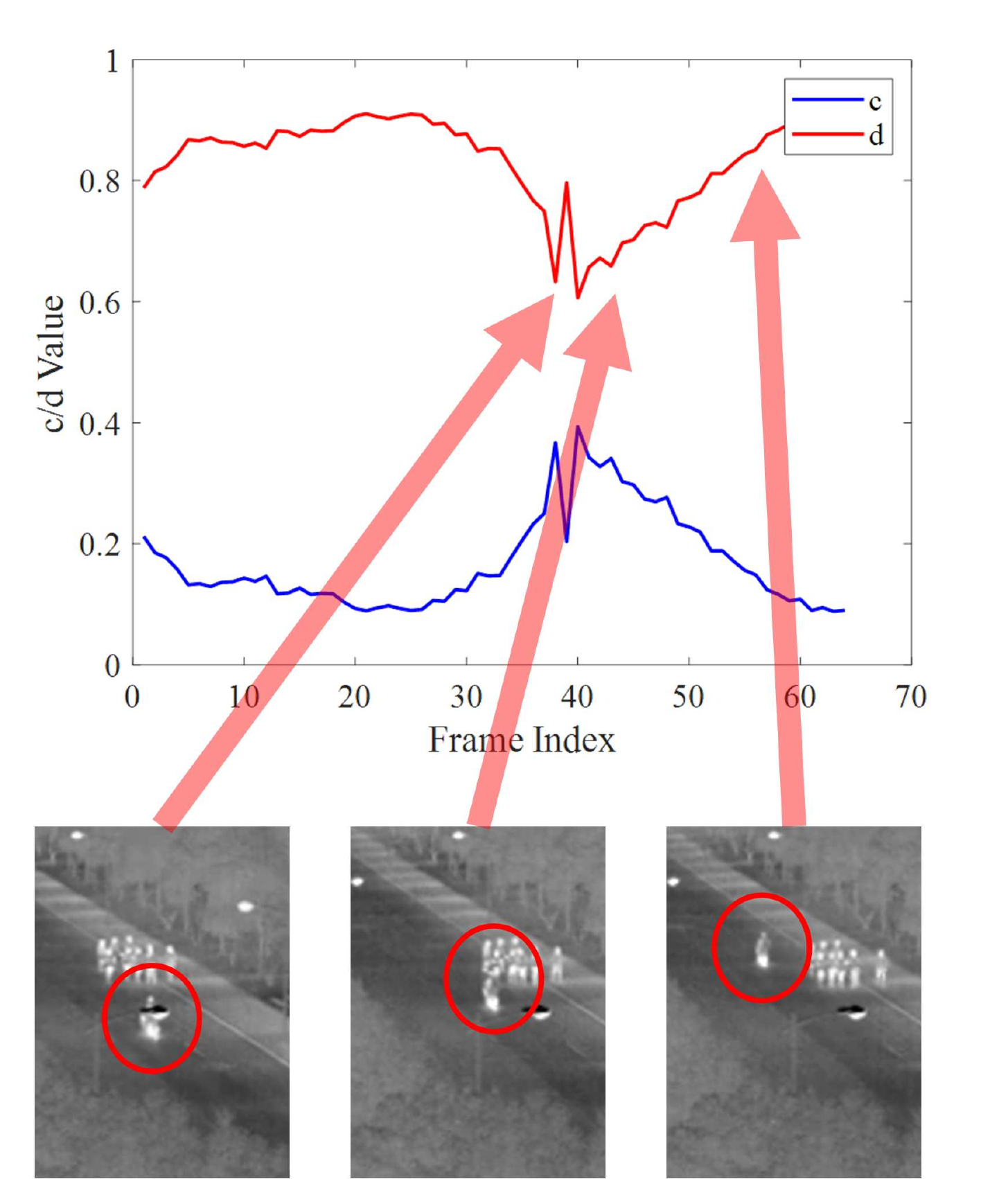}
\par\end{centering}
}
\par\end{centering}
\caption{\label{fig:7}Dynamic fusion layer weights of different frames of
(a) \textit{OccCar-2 }and (b) \textit{FastMotorNig}}
\end{figure}

In addition, we visualized the weights of the dynamic fusion layer
in two video sequences, as shown in Figure \ref{fig:7}. Figure \ref{fig:7}
(a) is from \textit{OccCar-2}. It can be found that in this video
sequence, the contributions of individual features and common features
are also different. At the beginning of the video, the car is clear
in the visible images, and the contributions of individual features
of visible images are large. When the car is blocked by leaves, visible
images cannot clearly distinguish the car, so the contributions of
the individual features reduce, while the contributions of the common
features increase. As the car comes out of the leaves, the contributions
of individual features of visible images increase. Figure \ref{fig:7}
(b) is from \textit{FastMotorNig}. When the bicycle is blocked by
a street light, the contributions of the individual features reduce,
while the contributions of the common features increase. In the later
stage of the video sequence, because the bicycle is close to the crowd,
\textit{thermal crossover} \cite{ref23} occurs, so the contributions
of the individual features of the infrared images are low. As the
bicycle moves away from the crowd, the contributions of the individual
features of the infrared images increase.

\subsection{Efficiency Analysis\label{subsec:Efficiency-Analysis}different fusion
methods}

\begin{figure}
\subfloat[Comparison of precision rate and speed]{\begin{centering}
\includegraphics[scale=0.4]{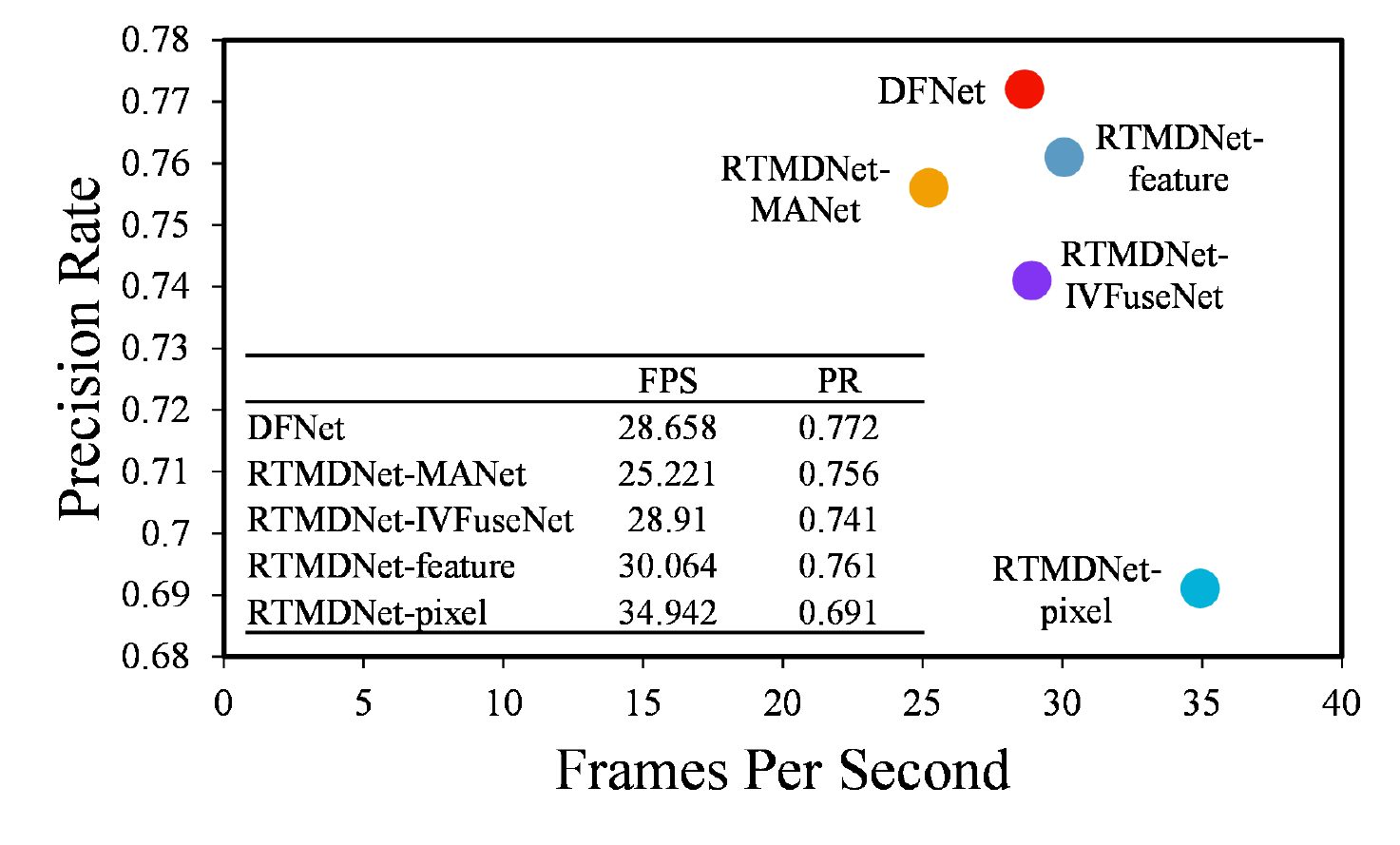}
\par\end{centering}
}\subfloat[Comparison of success rate and speed]{\begin{centering}
\includegraphics[scale=0.4]{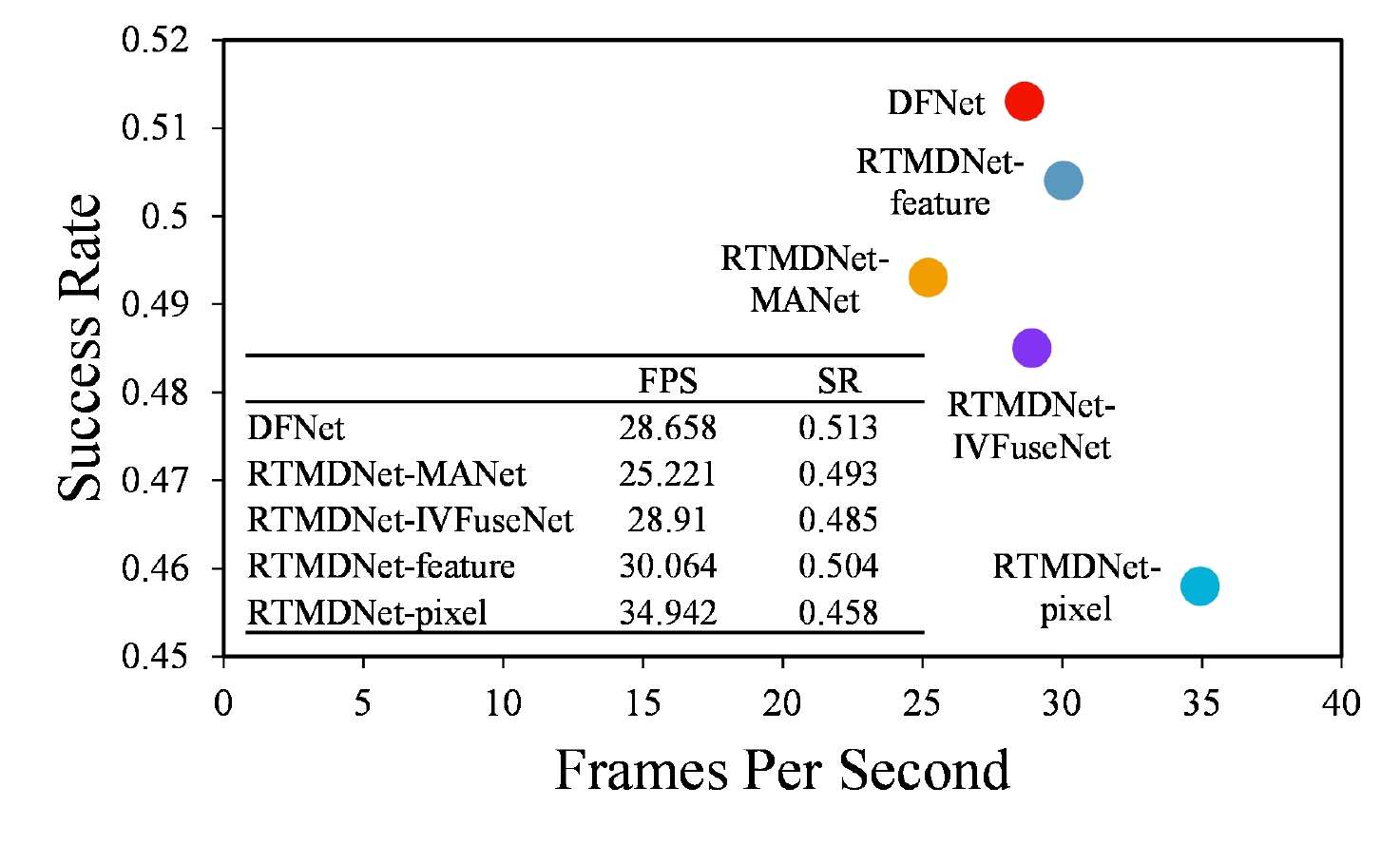}
\par\end{centering}
}

\caption{\label{fig:8}Comparison of speed and PR/SR.}
\end{figure}

The speed of DFNet is 28.658 FPS. We compared the speed and performance
of DFNet with other fusion tracking methods, the results are shown
in Figure \ref{fig:8}. The computational cost of DFNet is $O(X)=2(HWC_{in}+C_{in}C_{hidden}+2C_{hidden}+HWC_{in}C_{out}k^{2})$
Mult-Adds, where $H$, $W$ are the height and width of the input.
$C_{in}$, $C_{out}$, and $C_{hidden}$ are the channel numbers of
the input, output, and hidden layer, respectively. $k$ is the size
of the convolution kernel. Correspondingly, the computational cost
of baseline (RTMDNet-feature) is $O(X)=2HWC_{in}C_{out}k^{2}$ Mult-Adds.
Since the fusion of shared and non-shared convolution kernels is performed
in convolution kernel space, compared with the non-shared-convolution-kernel-based
fusion method, no additional calculations to increase. The increase
of computational cost is mainly due to the attention which calculates
weights according to the input. The computational cost caused by the
attention is much smaller than convolution. In DFNet, it is less than
0.02\%. While, compared with the baseline, MANet fuses the shared
and non-shared features in feature space, which causes calculations
to increase by 8.85\%. The specific computational cost is shown in
the Table \ref{tab:4}.

\begin{table}
\caption{\label{tab:4}The computational cost of different fusion methods.
C1, C2, C3, and total indicate the computational cost of the first,
second, third, and all convolutional layers, respectively. Percent
indicate computational cost expressed as a percentage of the non-shared-convolution-kernel-based
fusion method.}

\selectlanguage{british}%
\centering{}\resizebox{\textwidth}{!}{
\begin{tabular}{cccccc}
\hline 
\selectlanguage{english}%
Model\selectlanguage{british}%
 & \selectlanguage{english}%
C1\selectlanguage{british}%
 & \selectlanguage{english}%
C2\selectlanguage{british}%
 & \selectlanguage{english}%
C3\selectlanguage{british}%
 & \selectlanguage{english}%
total\selectlanguage{british}%
 & \selectlanguage{english}%
percent\selectlanguage{british}%
\tabularnewline
\hline 
\selectlanguage{english}%
RTMDNet-feature (baseline)\selectlanguage{british}%
 & \selectlanguage{english}%
323.14M\selectlanguage{british}%
 & \selectlanguage{english}%
768.00M\selectlanguage{british}%
 & \selectlanguage{english}%
285.47M\selectlanguage{british}%
 & \selectlanguage{english}%
1376.61M\selectlanguage{british}%
 & \selectlanguage{english}%
100.00\%\selectlanguage{british}%
\tabularnewline
\selectlanguage{english}%
RTMDNet-IVFuseNet\selectlanguage{british}%
 & \selectlanguage{english}%
323.14M\selectlanguage{british}%
 & \selectlanguage{english}%
768.00M\selectlanguage{british}%
 & \selectlanguage{english}%
285.47M\selectlanguage{british}%
 & \selectlanguage{english}%
1376.61M\selectlanguage{british}%
 & \selectlanguage{english}%
100.00\%\selectlanguage{british}%
\tabularnewline
\selectlanguage{english}%
RTMDNet-MANet\selectlanguage{british}%
 & \selectlanguage{english}%
382.49M\selectlanguage{british}%
 & \selectlanguage{english}%
798.72M\selectlanguage{british}%
 & \selectlanguage{english}%
317.19M\selectlanguage{british}%
 & \selectlanguage{english}%
1498.40M\selectlanguage{british}%
 & \selectlanguage{english}%
108.85\%\selectlanguage{british}%
\tabularnewline
\selectlanguage{english}%
DFNet (ours)\selectlanguage{british}%
 & \selectlanguage{english}%
323.21M\selectlanguage{british}%
 & \selectlanguage{english}%
768.12M\selectlanguage{british}%
 & \selectlanguage{english}%
285.57M\selectlanguage{british}%
 & \selectlanguage{english}%
1376.90M\selectlanguage{british}%
 & \selectlanguage{english}%
100.02\%\selectlanguage{british}%
\tabularnewline
\hline 
\end{tabular}}\selectlanguage{english}%
\end{table}

Based on all the experiments performed in this section, we conclude
that:
\begin{enumerate}
\item Compared with the visible tracking method (MDNet, RT-MDNet, SiamFC,
and SiamRPN) and the fusion tracking method (pixel-level fusion, feature-level
fusion, MANet, and IVFuseNet), DFNet achieves the best PR and SR.
\item The performance of fusion method is better than that of methods based
on single-modal images, which shows the advantage of fusion.
\item With consideration of the contributions of individual features and
common features, DFNet can adaptively calculate the weights of shared
and non-shared convolution kernels to cope with changes in modality
reliability.
\item Compared with the fusion of shared and non-shared features in the
feature space, the fusion of shared and non-shared convolution kernels
in the convolution kernel space can effectively reduce the computational
complexity and improve the tracking speed.
\end{enumerate}

\section{Conclusion}

In this paper, we propose a novel RGBT tracking method, called dynamic
fusion network (DFNet). DFNet is essentially a feature-level fusion
method, which can use non-shared convolutions to respectively extract
individual features according to the different characteristics of
visible and infrared images. Furthermore, DFNet takes the advantage
of shared convolution kernels to extract common features. In addition,
because attention is used to adaptively calculate different convolution
kernel weights according to inputs, DFNet can dynamically calculate
the contributions of individual features and common features in the
face of changes in modality reliability. The shared convolution kernels
and non-shared convolution kernels are concatenated in convolution
kernel space, so that, the computational cost is small. Extensive
experiments on two RGBT datasets validate the effectiveness of DFNet.
Future work will focus on adopting more advanced architectures, designing
other adaptive weighting methods, and reducing the redundancy of features
between different modalities.

\bibliographystyle{unsrt}
\bibliography{export}

\end{document}